%% file: entrance.tex
\documentclass[acmtog]{acmart}
\acmSubmissionID{934}

\usepackage{float}
\usepackage{multirow}
\usepackage{booktabs} % For formal tables

\usepackage[ruled]{algorithm2e} % For algorithms

\newcommand{\final}[1]{#1}
\newcommand{\MethodData}{\textsc{Mobjaverse\xspace}}
\newcommand{\MethodMotion}{\textsc{TopoCap\xspace}}

\SetAlFnt{\small}
\SetAlCapFnt{\small}
\SetAlCapNameFnt{\small}
\SetAlCapHSkip{0pt}

\acmJournal{TOG}
\copyrightyear{2026}
\acmYear{2026}
\setcopyright{cc}
\setcctype{by}
\acmConference[SIGGRAPH Conference Papers '26]{Special Interest Group on Computer Graphics and Interactive Techniques Conference Conference Papers}{July 19--23, 2026}{Los Angeles, CA, USA}
\acmBooktitle{Special Interest Group on Computer Graphics and Interactive Techniques Conference Conference Papers (SIGGRAPH Conference Papers '26), July 19--23, 2026, Los Angeles, CA, USA}
\acmDOI{10.1145/3799902.3811159}
\acmISBN{979-8-4007-2554-8/2026/07}

\begin{document}
% Title portion
% \title{\MethodMotion: Learning Topology-Agnostic Motion Priors for Monocular Video-to-Animation}
\title{\MethodMotion: Learning Topology-Agnostic Motion Priors for Monocular Video-to-Animation}

% DO NOT ENTER AUTHOR INFORMATION FOR ANONYMOUS TECHNICAL PAPER SUBMISSIONS TO SIGGRAPH 2019!
\author{Cheng-feng Pu}
\orcid{0009-0005-4483-2751}
\affiliation{%
 \institution{Zhili College, Tsinghua University}
 \country{China}
}
\email{pcf22@mails.tsinghua.edu.cn}

\author{Jia-peng Zhang}
\orcid{0009-0001-9502-9484}
\affiliation{
 \institution{BNRist, Department of Computer Science and Technology, Tsinghua University}
 \country{China}
}
\email{zjp24@mails.tsinghua.edu.cn}

\author{Meng-hao Guo}
\orcid{0000-0002-4128-4594}
\affiliation{
 \institution{BNRist, Department of Computer Science and Technology, Tsinghua University}
 \country{China}
}
\email{gmh@tsinghua.edu.cn}

\author{Yan-Pei Cao}
\orcid{0000-0002-0416-4374}
\affiliation{
 \institution{VAST}
 \country{China}
}
\email{caoyanpei@gmail.com}

\author{Shi-Min Hu}
\authornote{Corresponding author.}
\orcid{0000-0001-7507-6542}
\affiliation{
 \institution{BNRist, Department of Computer Science and Technology, Tsinghua University}
 \country{China}
}
\email{shimin@tsinghua.edu.cn}

\begin{abstract}

The explosion of generative 3D assets has created a massive demand for animation, yet current motion capture methods remain brittle, restricted to species-specific templates (e.g., SMPL) or requiring labor-intensive manual rigging.
We introduce \MethodMotion, the first unified framework capable of \final{extracting} motion from monocular video and retargeting it onto characters with arbitrary, unseen skeletal topologies, i.e., from bipeds to hexapods and inanimate objects, without test-time optimization.
Our key insight is that while skeletal structures are combinatorial and discrete, the underlying physics of motion occupy a continuous, low-dimensional manifold.
We materialize this insight via a two-stage generative pipeline. First, we learn a {Universal Motion Manifold} using a Graph CVAE that compresses heterogeneous kinematic chains into a shared, fixed-length latent code. By explicitly conditioning the decoder on a structural embedding of the target rig, we disentangle motion dynamics from skeletal topology.
Second, we treat video-to-animation as a conditional flow matching problem, predicting these topology-agnostic codes from visual features.
To learn this generalized prior, we introduce \MethodData, a massive-scale dataset curated from Objaverse-XL. Comprising over \final{5,000} unique skeletal topologies and \final{2} million frames, it exceeds the structural diversity of existing datasets by two orders of magnitude.
Extensive experiments demonstrate that \MethodMotion outperforms specialist models on human and quadruped benchmarks while enabling zero-shot retargeting for the long tail of 3D creatures. \final{Dataset is publicly available at https://huggingface.co/datasets/duckduckplz/Mobjaverse}
\end{abstract}

\begin{teaserfigure}
\centering
  \includegraphics[width=0.95\textwidth]{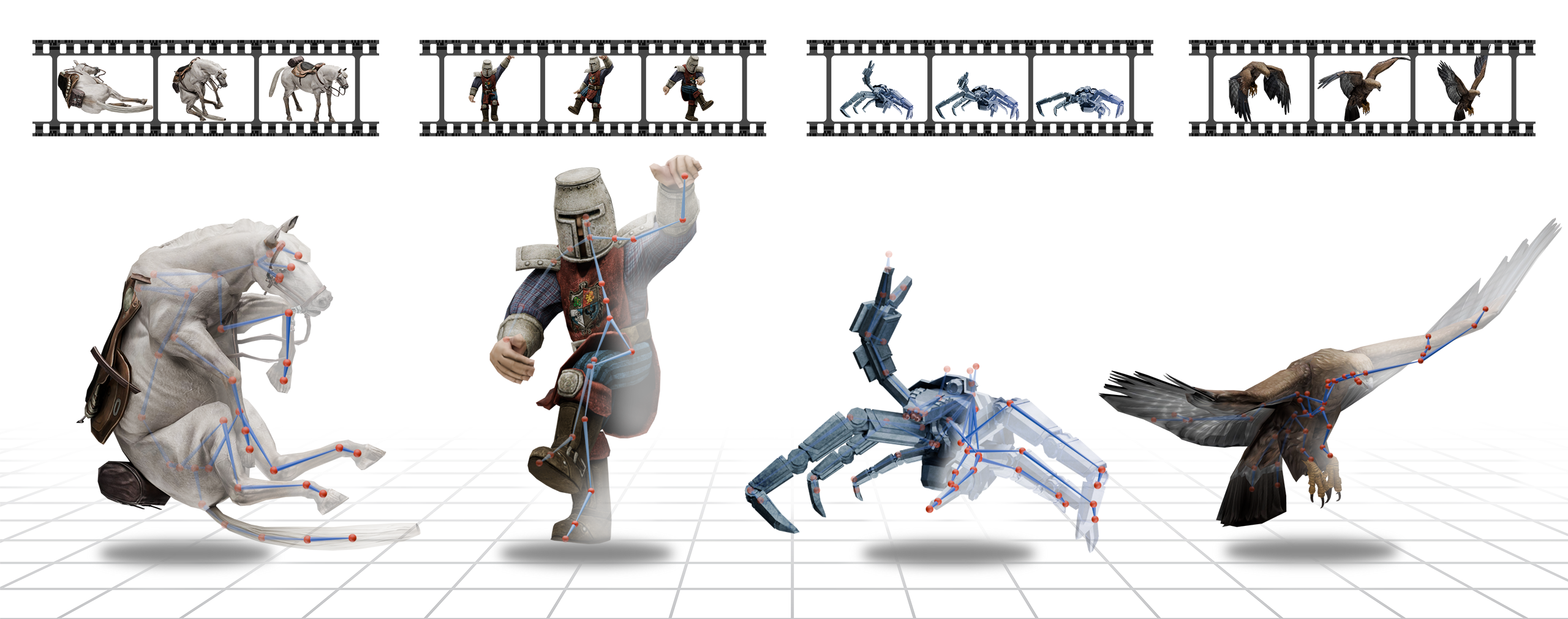}
  \caption{\textbf{\MethodMotion: \final{Universal Motion Priors for Video-Driven 3D Animation.}} We introduce the first topology-agnostic framework capable of extracting motion from video and retargeting it onto \textit{arbitrary} 3D characters in a zero-shot manner. Our method learns a unified motion manifold that generalizes across diverse morphologies (bipeds, quadrupeds, hexapods, and flying creatures) without requiring template priors or test-time optimization.}
  \label{fig:teaser}
\end{teaserfigure}

\begin{CCSXML}
<ccs2012>
<concept>
<concept_id>10010147.10010371.10010352.10010380</concept_id>
<concept_desc>Computing methodologies~Motion processing</concept_desc>
<concept_significance>500</concept_significance>
</concept>
<concept>
<concept_id>10010147.10010178.10010224.10010240.10010242</concept_id>
<concept_desc>Computing methodologies~Shape representations</concept_desc>
<concept_significance>500</concept_significance>
</concept>
</ccs2012>
\end{CCSXML}

\ccsdesc[500]{Computing methodologies~Motion processing}
\ccsdesc[500]{Computing methodologies~Shape representations}

%
% End generated code
%

\keywords{Topology-Agnostic Animation, Universal Motion Priors, Generative Motion Capture, Zero-Shot Retargeting}

\maketitle

\input{sections/intro}
\input{sections/related_work}

\input{sections/dataset}
\input{sections/method}
\input{experiments_and_discussion}

\end{document}

%% file: sections/intro.tex
\section{Introduction}

The rapid advancement of generative AI has democratized the creation of static 3D geometry~\citep{triposg,hunyuan3d,peng2024charactergen,lxl,diffusionmodels}. However, animating this expanding universe of digital assets remains a significant bottleneck.
While monocular motion capture has achieved high fidelity for standardized subjects like humans~\citep{fusion4d, kocabas2020, rempe2021} and quadrupeds~\citep{xie2025animamimicimitating3danimation, AnimalAvatars2024}, these methods rely on parametric templates (e.g., SMPL~\citep{smpl}, MHR~\citep{mhr}, SMAL~\citep{smal}) that encode strong anatomical priors. 
Such template-based paradigms are fundamentally non-scalable: they fail catastrophically when applied to the \textit{long tail} of 3D content, from fantasy creatures to articulated furniture, where no pre-defined template exists.

A natural alternative is motion retargeting, which aims to transfer motion between different skeletons. Classical retargeting methods ~\citep{deepmotionretarget, posetomotion, motion2motion} typically assume shared semantics or similar skeletal structures, and often rely on careful normalization or optimization. While effective within limited domains, these approaches remain constrained by structual compatibility and do not generalize to the open-world setting with drastically different topologies.

The core technical barrier is the rigid coupling of \textit{motion} and \textit{structure}. 
Traditionally, motion is represented as a sequence of local joint rotations relative to a specific kinematic hierarchy, making data mathematically incompatible across skeletons. 
To animate the infinite variety of generated 3D characters, one would theoretically require a distinct motion prior for every possible skeletal topology—an intractable proposition.
To date, no framework can extract high-fidelity motion from video and map it directly onto an arbitrary target skeleton without extensive manual optimization.

In this work, we propose \textbf{\MethodMotion}, a framework that breaks this barrier by learning a \textit{Universal Motion Representation}. 
Our central hypothesis is that while skeletal morphologies vary infinitely, the underlying dynamics of locomotion and gesture share a common, low-dimensional intrinsic space. 
If we can map diverse kinematic structures into this shared manifold, the problem of motion capture transforms from specific pose estimation to general motion synthesis.

We realize this through a two-stage disentangled generative pipeline. First, we learn a topology-agnostic motion manifold using a Perceiver-based CVAE~\citep{jaegle2021perceiverio}. By conditioning the CVAE on the target rest pose, we force the \textit{fixed-length} latent code to capture only structure-invariant dynamics (what the character is \textit{doing}) while the decoder handles the kinematic execution (how the body \textit{moves}). Second, a Latent Flow Matching model predicts these latent codes from DINOv3~\citep{simeoni2025dinov3} video features. By predicting in this shared latent space rather than the complex joint space, our model robustly aligns visual cues with arbitrary structures, enabling zero-shot transfer.

A data-driven universal prior requires structural diversity that no existing dataset provides. 
Current benchmarks are either massive but topologically homogeneous (human-centric~\citep{humanml3d, fan2025zerozeroshotmotiongeneration, lafan, amass, human36m}) or diverse but small~\citep{wang2025animo, truebones, yang2023omnimotiongpt, zoo300k}. 
To bridge this gap, we introduce \textbf{\MethodData}, the largest structurally diverse motion dataset to date. Mined from Objaverse-XL~\citep{objaverseXL}, we apply a rigorous pipeline of kinematic validation, standardization, and semantic filtering to curate $5,006$ distinct skeletal topologies, providing the critical mass of variation needed to learn a topology-agnostic prior. Our framework is trained and quantitatively benchmarked on synthetically rendered videos derived from this dataset, and the learned priors demonstrate promising zero-shot generalization to real-world internet videos.

In summary, our contributions are:
\begin{enumerate}
    \item{\textbf{A Universal Motion Representation:} We propose a novel Graph CVAE architecture that disentangles motion dynamics from skeletal structure, mapping variable kinematic chains to a shared, fixed-length latent space.}
    \item{\textbf{\MethodMotion:} We present the first generative framework capable of extracting motion from monocular video for arbitrary skeletal topologies in a single forward pass, eliminating the dependency on parametric templates.}
    \item{\textbf{\MethodData:} We release a massive-scale motion dataset containing thousands of unique topologies, exceeding the structural diversity of prior art by orders of magnitude, to facilitate future research in generalist animation.}
\end{enumerate}

%% file: sections/related_work.tex
\vspace{-2.5mm}
\section{Related Work}

\subsection{Motion Capture with Priors}
Recovering 3D motion from visual observations is ill-posed, necessitating strong priors. \textit{Template-based learning methods} resolve ambiguity by encoding motion through parametric models for humans~\citep{smpl, smplx, kocabas2020, dong2020, rempe2021} or animals~\citep{smal, bird, lassie, magicpony}. While robust within their domains, these priors are strictly bound to fixed skeletal structures, failing to generalize to the "long tail" of arbitrary topologies found in the wild.

\textit{Topology-agnostic approaches} attempt to bypass templates by introducing alternative priors. Some leverage 3D mesh generative priors~\citep{mocapanything} or exploit marker or flow-based tracking~\citep{Chen2022LearningVM, song2025puppeteer, xie2025animamimicimitating3danimation}. Others incorporate high-level semantic guidance from text-to-image models~\citep{deb2025articulate3dzeroshottextdriven3d} or video foundation models~\citep{li2025articulatedkinematicsdistillationvideo}. Despite reducing template dependence, these methods typically rely on computationally expensive 4D mesh reconstruction or iterative optimization (e.g., IK), limiting scalability.
\textit{Physics-based methods}~\citep{amp, ppmotion, deepmimic, skillmimicv2} ensure plausibility via simulation but are sensitive to modeling inaccuracies and reward design. 
\textit{In contrast, our method learns a unified, topology-agnostic kinematic prior that aligns directly with video latent features, enabling efficient motion capture without fixed templates, handcrafted physical objectives, or expensive 4D reconstruction.}

\subsection{Motion Generation}
Generative motion modeling has advanced rapidly via diffusion~\citep{xmogen, zhang2025towards, yang2023omnimotiongpt, mixermdm, lgtm, energymogen} and autoregressive architectures~\citep{zhong2023attt2m, zhang2023generating, pinyoanuntapong2024mmm, atom, smoogpt, liao2025shape}. However, this success is largely predicated on \textit{unified parametric templates} (e.g., SMPL/SMAL), which allow models to leverage homogeneous human\final{~\citep{humanml3d, fan2025zerozeroshotmotiongeneration, chen2023executing, tevet2023human, human36m, lafan, amass, Hou2024}} or quadruped~\citep{wang2025animo, zoo300k, truebones} datasets.

This reliance on fixed templates fundamentally limits expressiveness for non-standard characters. While recent efforts have explored generation across varying topologies~\citep{anytop, raab2024single, ganimator, animax}, they are bottlenecked by the scarcity of large-scale, topology-diverse motion data.
\textit{We address this by explicitly disentangling motion dynamics from skeletal structure and introducing \MethodData, the largest structurally diverse motion dataset to date, enabling universal motion synthesis that transcends fixed parametric templates.}

%% file: sections/dataset.tex
\section{\MethodData: A Foundation for Generalist Motion}

A key bottleneck in learning topology-agnostic priors is the structural ``polarization'' of existing motion data.Current benchmarks are either high-quality but topologically homogeneous (human-centric~\citep{amass, humanml3d, motionx, lafan, human36m} or quadruped-centric~\citep{wang2025animo, truebones, zoo300k}), or geometrically diverse but lacking in articulation (static 3D objects). This lack of structural diversity, especially the long tail of non-standard kinematic chains, limits learning a universal motion manifold. To bridge this gap, we introduce \textbf{\MethodData}, built from Objaverse-XL~\citep{objaverseXL} via a rigorous curation pipeline that extracts physically plausible kinematic chains from unconstrained web-scraped assets.

\begin{figure}[h]
    \centering
    \includegraphics[width=0.9\linewidth]{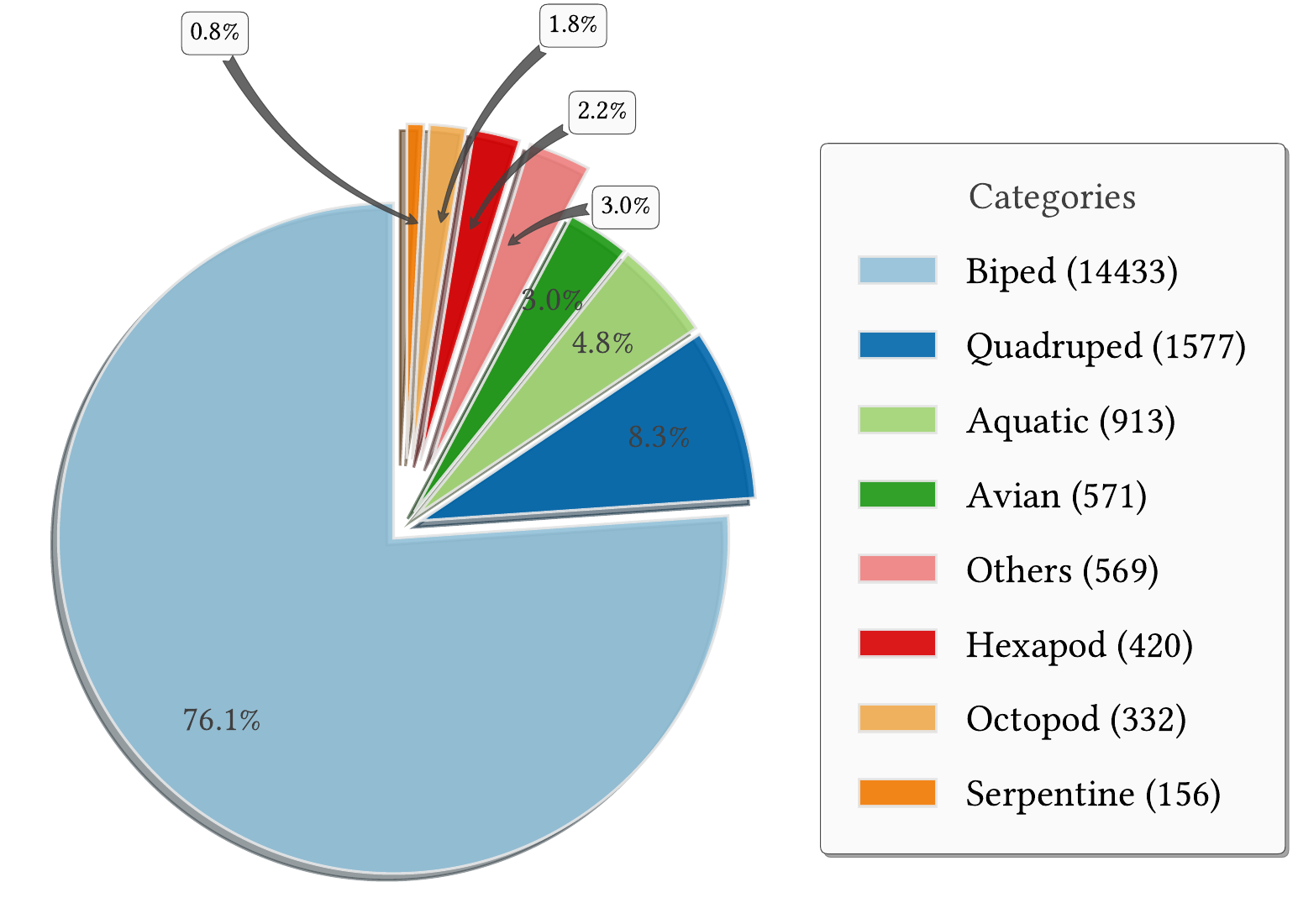}
    \caption{\textbf{Label distribution in \MethodData.}
    While Bipeds and Quadrupeds are prominent, \MethodData~contains a heavy tail of diverse topologies (Hexapods, Arachnids, Furniture) absent in previous datasets. This structural variance is critical for learning generalist priors.
    }
    \label{fig:category_distribution}
\end{figure}

\subsection{Curation Pipeline}
Raw Internet assets often suffer from broken hierarchies, degeneracy, or lack of meaningful motion. We implement a \final{five-stage} filtration pipeline to ensure geometric and semantic validity (see Supplementary A. for specific heuristics):

\paragraph{1. Kinematic Tree Validation.}
We first filter for valid hierarchical graph structures, discarding disjoint skeletons or cyclic dependencies. We retain kinematic trees with joint counts $J \in [2,128]$, a range capturing everything from simple hinged objects to complex multi-legged creatures while filtering out noise (e.g., single-bone props) and overly dense automated rigs.

\paragraph{2. Motion Standardization.}
We unify the training space by normalizing the global scale (unit bounding box) and realigning root trajectories. Crucially, we detect and collapse \textit{pseudo-root} chains, i.e., redundant nodes introduced by file exporters, and prune leaf nodes with zero skinning weights. This ensures the graph topology strictly reflects the effective surface deformation. \final{Also we merge temporally redundant frames, i.e., consecutive frames with negligible motion, to remove static segments and ensure that all retained sequences exhibit perceptually meaningful motion.}

\paragraph{3. VLM-Based Semantic Filtering.}
A significant portion of assets contain only very subtle animations. Naive variance thresholds fail here. We employ a Vision-Language Model (GPT-5.2~\citep{gpt52}) as a semantic discriminator. We query the VLM to classify rendered views as ``Dynamic and Meaningful'' or ``Static or Broken,'' filtering out thousands of degenerate samples.

\paragraph{4. Manual Verification.}
Automated metrics cannot fully capture surface artifacts like mesh tearing or ``candy-wrapper'' twisting. We perform a final expert manual verification step to discard assets with severe skinning errors or topologically implausible structures.

\paragraph{5. Texture Binding.}
To improve visual diversity, we enrich assets with missing or low-quality textures using Tripo3.0~\citep{triposg}. This augmentation introduces more realistic appearance variations, enabling the model to better generalize to complex scenarios.

\subsection{Dataset Statistics}
The resulting \textsc{Mobjaverse} contains \textbf{\final{5,006} unique skeletal topologies} and over \textbf{\final{2} million frames} of animation. This exceeds the topological diversity of existing animal datasets by two orders of magnitude. Crucially, as shown in Fig.~\ref{fig:category_distribution}, it covers a continuous spectrum of morphology, including hexapods, arachnids, and non-biological articulated objects (e.g., animated furniture, robots). This structural variance is the critical enabler for our method's ability to disentangle motion dynamics from topology.

%% file: sections/method.tex
\section{Method}
\label{sec:method}

\subsection{Preliminaries \& Problem Formulation}
\label{sec:preliminaries}
We formulate generalist motion capture as learning a conditional mapping from \final{a video sequence} to the kinematic state space of an arbitrary, user-specified 3D skeleton. 
Let $\mathcal{V} = \{ I_t \}_{t=1}^{T}$ denote an input sequence of $T$ RGB images. Let the target asset be defined by a tuple $\mathcal{A} = (\mathcal{M}, \mathcal{S})$, where $\mathcal{M}$ represents the surface mesh and $\mathcal{S}$ represents the skeletal rig. Our objective is to synthesize a motion sequence $\mathcal{B} = \{ \boldsymbol{B}_t \}_{t=1}^{T}$ that, when applied to $\mathcal{S}$, induces a mesh deformation $\mathcal{M}(t)$ that faithfully reconstructs the dynamics observed in $\mathcal{V}$.

\paragraph{Skeletal Topology.}
We define the skeleton $\mathcal{S}$ as a directed acyclic graph (tree) comprising $J$ joints. The topology is characterized by its connectivity (parent indices $f(\cdot)$) and rest-pose configuration. For each joint $i$, the rest-pose transformation relative to its parent is given by $M_i = (p_i, q_i)$, where $p_i \in \mathbb{R}^3$ is the rest offset and $q_i \in \mathbb{S}^3$ is the rest orientation quaternion.

\paragraph{Universal Motion Representation.}
Standard parametric representations (e.g., SMPL pose parameters) are topologically brittle. To animate diverse characters ranging from rigid robots to squash-and-stretch cartoons, we require a representation that is elastic and topology-agnostic. We parameterize the motion state $\boldsymbol{B}_t = (\boldsymbol{R}_t, \boldsymbol{O}_t)$ at frame $t$ as:
\begin{itemize}
    \item \textbf{Local Rotations} $\boldsymbol{R}_t = \{r_{t,i} \in \mathbb{S}^3\}_{i=1}^J$: The relative rotation of joint $i$ with respect to its parent frame.
    \item \textbf{Elastic Offsets} $\boldsymbol{O}_t = \{o_{t,i} \in \mathbb{R}^3\}_{i=1}^J$: Time-varying translational deltas added to the rest bone lengths. This term captures non-rigid structural deformations (e.g., breathing, cartoon elasticity) that cannot be modeled by rotation alone.
\end{itemize}
The \textit{global} pose of joint $i$ at time $t$, denoted $G_{t,i} = (p^*_{t,i}, q^*_{t,i})$, is derived via a differentiable Forward Kinematics (FK) operator. Unlike standard rigid-body FK, our operator $\text{FK}(\boldsymbol{R}_t, \boldsymbol{O}_t, \mathcal{S})$ incorporates the elastic offsets $\boldsymbol{O}_t$, allowing the character morphology to adapt dynamically to the visual evidence.

\begin{figure*}
    \centering
    \includegraphics[width=0.9\linewidth]{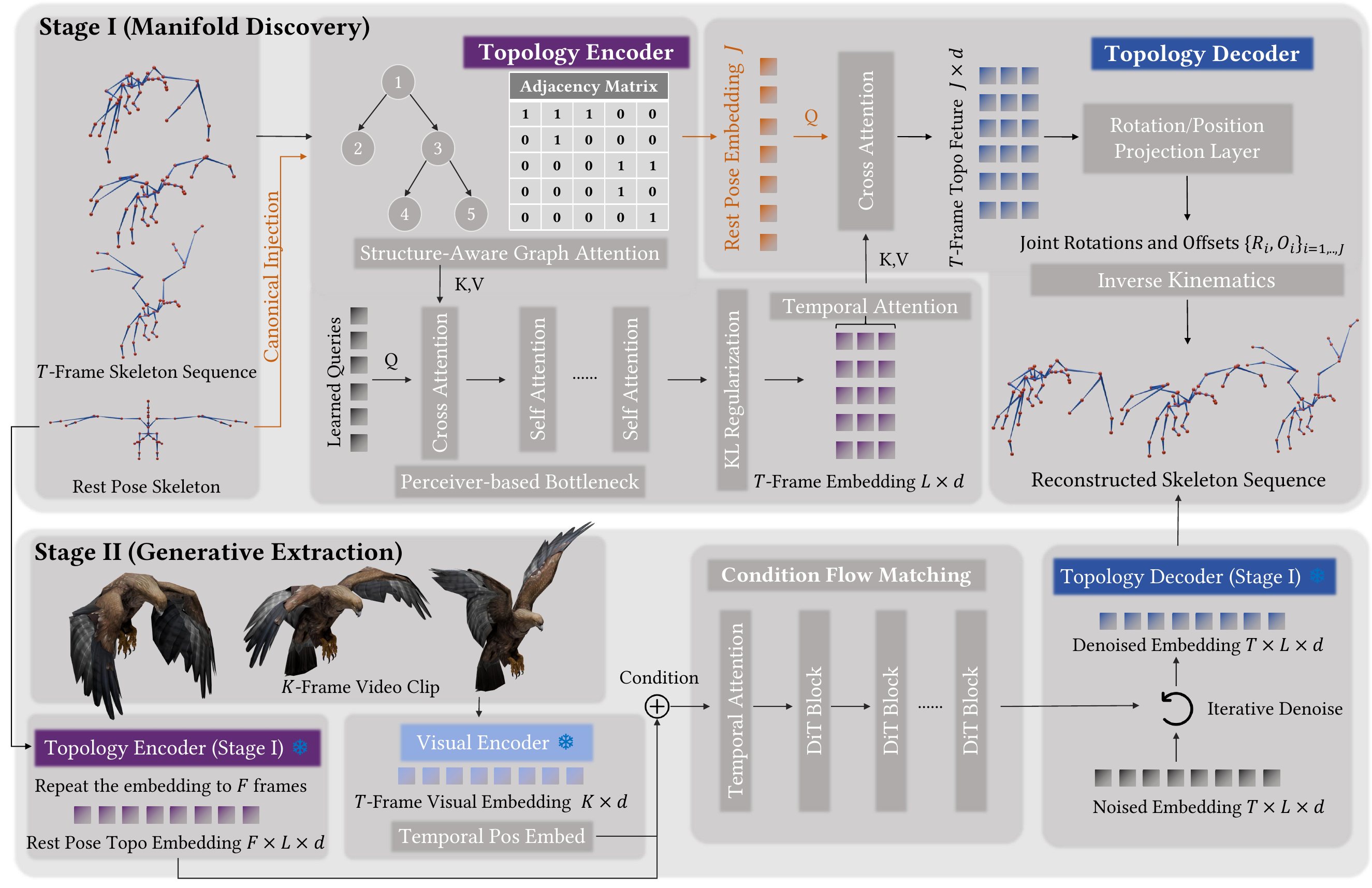}
    \caption{\textbf{Overview of \MethodMotion.} 
    The framework operates via a two-stage generative pipeline. \textbf{Stage I (Manifold Discovery):} A Graph CVAE compresses motion from heterogeneous skeletons into a shared, fixed-length latent manifold ($K\times D$) using a Perceiver-based bottleneck. A topology-conditioned decoder reconstructs the motion using analytic Inverse Kinematics (IK) to ensure global consistency. \textbf{Stage II (Generative Extraction):} We treat motion capture as a conditional flow matching problem. A frozen visual encoder extracts video features, which are fused with a structural embedding of the target rig (via \textit{Canonical Injection}). The flow transformer predicts the latent motion code, which is then decoded by the Stage-I decoder to produce the final animation.
    }

    \label{fig:pipeline}
\end{figure*}

\subsection{\MethodMotion~Framework Overview}
\label{sec:overview}
The core hypothesis of \MethodMotion~is that while skeletal topologies are discrete and combinatorial, the manifold of physically plausible motions is continuous and low-dimensional. Direct regression from pixels to arbitrary graph structures is ill-posed due to the lack of a shared output space. To resolve this, we decouple \textit{motion dynamics} from \textit{skeletal structure} through a two-stage generative pipeline (Fig.~\ref{fig:pipeline}):
\begin{enumerate}
    \item \textbf{Manifold Discovery (Sec.~\ref{sec:cvae}):} We first learn a universal motion prior by training a topology-agnostic Conditional Variational Autoencoder (CVAE)~\citep{sohn2015learning}. This model compresses motion from heterogeneous skeletons into a shared, fixed-length latent code $z \in \mathbb{R}^{L \times D}$. By conditioning the reconstruction on the explicit rig structure $\mathcal{S}$, we force the latent code $z$ to capture structure-invariant dynamics rather than joint-specific coordinates.
    \item \textbf{Generative Extraction (Sec.~\ref{sec:diffusion}):} Utilizing the learned manifold, we reformulate video-to-animation as a conditional generation task. We train a Flow Matching model to predict motion codes $z$ from visual features $\mathcal{V}$, conditioned on the target rig $\mathcal{S}$. This allows us to navigate the latent space to synthesize plausible motion for unseen skeletons.
\end{enumerate}

\subsection{Stage I: Learning a Universal Motion Manifold}
\label{sec:cvae}
To construct a unified representation space for disparate skeletons, we implement a Graph CVAE~\citep{sohn2015learning}. The fundamental challenge is that conventional neural networks require fixed-dimension inputs, whereas skeletal graphs vary in node count $J$ and connectivity. We resolve this via a \final{\textit{compress-then-compose}} architecture.
\subsubsection{Structure-Aware Motion Encoding}
\label{sec:structureaware}
Given a motion sequence $(\boldsymbol{R}, \boldsymbol{O})$ on skeleton $\mathcal{S}$, we first embed raw kinematic data into high-dimensional joint features. For a specific frame, let $\mathbf{h}_i^{(0)}$ denote the initial embedding of joint $i$, formed by concatenating the local rotation $r_{i}$, elastic offset $o_{i}$, and rest-pose position $p_i$.

To capture kinematic dependencies without overfitting to specific topologies, we employ a graph-based multi-head attention mechanism~\citep{velickovic2018graph}. Unlike global self-attention, we restrict information flow to immediate neighbors (parents and children) to enforce kinematic causality. The updated feature for joint $i$ at layer $l$ is:
\begin{equation}
    % \mathbf{h}_i^{(l+1)} \leftarrow \sum_{j=1}^{J} E_{i,j} \frac{\left(\boldsymbol{Q}\mathbf{h}_i^{(l)}\right)^\top \left(\boldsymbol{K}\mathbf{h}_j^{(l)}\right)}{\sqrt{d_{head}}} \boldsymbol{V}\mathbf{h}_j^{(l)} + \mathbf{P}_i,
    \mathbf{h}_i^{(l+1)} \leftarrow \sum_{j \in \mathcal{N}(i)} \frac{\left(\boldsymbol{Q}\mathbf{h}_i^{(l)}\right)^\top \left(\boldsymbol{K}\mathbf{h}_j^{(l)}\right)}{\sqrt{d_{\text{head}}}} \boldsymbol{V}\mathbf{h}_j^{(l)} + \mathbf{P}_i,
\end{equation}
where $\mathcal{N}(i)$ is the neighborhood of $i$, $\boldsymbol{Q}, \boldsymbol{K}, \boldsymbol{V}$ are learnable projection matrices, and $\mathbf{P}_i$ is a learnable embedding encoding the joint's structural role. We stack $N$ such blocks, allowing kinematic information to propagate iteratively through the chain while maintaining permutation invariance.

\subsubsection{Fixed-Length Latent Compression} 
\label{sec:perceiver}
The graph encoder outputs variable-length features $\mathbf{H} = \{\mathbf{h}_i\}_{i=1}^J$. To map this to a universal manifold, we employ a Perceiver-IO~\citep{jaegle2021perceiverio,zhang20233dshape2vecset,triposg} bottleneck.
We define $L$ learnable latent queries $\boldsymbol{\Lambda} \in \mathbb{R}^{L \times d}$, where $L$ is fixed independent of skeleton complexity. These queries extract abstract motion concepts via Cross-Attention:
\begin{equation}
    \mathbf{Z}^{(0)} = \text{Attention}(Q=\boldsymbol{\Lambda}, K=\mathbf{H}, V=\mathbf{H}).
\end{equation}
This bottleneck forces the model to compress specific joint movements into high-level dynamic descriptions. We then apply standard Transformer self-attention layers on $\mathbf{Z}$ to model correlations between these abstract concepts.
Crucially, this compression is spatial only; we preserve the temporal resolution $T$, yielding a latent distribution $z \in \mathbb{R}^{T \times L \times d}$. This preserves high-frequency motion details vital for crisp animation.

\subsubsection{Global-First Reconstruction with Analytic IK}
\label{sec:decoding}
The decoder $\mathcal{D}$ mirrors the encoder. To reconstruct motion for a target skeleton $\mathcal{S}$, we first embed $\mathcal{S}$ into structural queries $\{\mathbf{q}_i\}_{i=1}^J$ (Sec.~\ref{sec:structureaware}). These queries attend to the motion latents $z$ to retrieve dynamics relevant to each specific joint.

\paragraph{Analytic Inverse Kinematics}
Directly predicting local rotations often leads to error accumulation along deep kinematic chains. Conversely, predicting only global positions ignores skeletal constraints. We propose a \textit{global-first} reconstruction scheme. The decoder predicts \textit{global} joint positions $p^*_{t,i}$ and \textit{global} orientations $m^*_{t,i}$. We then recover the local pose parameters (deviations from rest) via differentiable Inverse Kinematics.
For joint $i$ with parent $f_i$, let $m_i$ denote its rest-pose local rotation. We recover the predicted local pose rotation $\hat{r}_{t,i}$ and elastic offset $\hat{o}_{t,i}$ as:
\begin{equation}
\begin{cases}
\begin{aligned}
\hat{r}_{t,i} &= m_i^{-1} m_{f_i} (m^*_{t,f_i})^{-1} m^*_{t,i},\\
\hat{o}_{t,i} &= m_i^{-1}\!\left(
m_{f_i}(m_{f_i}^*)^{-1}(p_{t,i}^*-p_{t,f_i}^*)-(p_i-p_{f_i}^{})\right),
\end{aligned}
\end{cases}
\end{equation}
where $m_{f_i}=m_{t,f_{i}}^*=I$ and $p_{f_i}=p_{t,f_i}^*=0$ for root joint. This formulation ensures that the final motion is globally consistent while maintaining a valid local parameterization for the skeletal graph. The detailed derivation can be found in Supplementary B.

\subsubsection{Training Objectives} 
The CVAE is trained end-to-end minimizing Mean Squared Error (MSE) on global positions and orientations. To resolve the double-cover ambiguity of quaternions, we minimize the cosine distance $1 - \langle m_{t,i}^*, \hat{m}_{t,i}^* \rangle^2$.
To ensure temporal smoothness, we impose an acceleration loss $\mathcal{L}_{acc}$~\citep{zeng2022smoothnet} on the predicted global positions. The latent space is regularized via the standard KL-divergence term $\mathcal{L}_{KL}$. See Supplementary C.2 for formal definitions.

\subsection{Stage II: Generative Motion Extraction}
\label{sec:diffusion}
With the universal motion manifold $\mathcal{Z}$ established, monocular motion capture becomes a trajectory finding problem: we seek a path $z_{1:T} \in \mathcal{Z}$ that aligns with visual evidence $\mathcal{V}$ and target topology $\mathcal{S}$. We model $p(z_{1:T} | \mathcal{V}, \mathcal{S})$ using a Flow Matching transformer~\citep{lipman2022flow}.

\subsubsection{Multi-Modal Conditioning}
\label{sec:conditioning}
To generalize to unseen characters, the model must understand the correlation between the visual appearance of a character from videos and its kinematic skeleton.

\paragraph{Visual Feature Extractor} 
We extract semantic motion cues using a frozen DINOv3~\citep{simeoni2025dinov3} encoder, $\phi_{\text{img}}$. DINOv3 provides consistent features across diverse object categories and viewpoints. We extract patch-level tokens for each frame $I_t$, insert learnable tokens for occluded or masked frames, and augment them with sinusoidal positional embeddings~\citep{attn} to form the visual context $\mathbf{C}_{\text{vis}}$.

\paragraph{Canonical Structural Injection} 
A critical design of our approach is how we inform the flow model about the target topology. Simply passing raw bone vectors is insufficient as it lacks semantic context. 
Instead, we introduce \textit{Canonical Structural Injection}. We utilize the pre-trained CVAE encoder $\mathcal{E}$ as a domain projector. Given the target rest-pose $\mathcal{S}$, we construct a ``zero-motion'' sequence (identity rotations, zero offsets) and pass it through $\mathcal{E}$ to obtain latent tokens $\mathbf{C}_{\text{skel}}$. 
This projects the target topology into the \textit{same latent manifold} as the motion targets. Consequently, the flow model can utilize cross-attention to establish a semantic correspondence between the abstract visual motion cues and the specific kinematic capabilities of the target rig.

\subsubsection{Spatiotemporal Flow Matching}
We employ a Diffusion Transformer (DiT) backbone adapted from Hunyuan3D 2.1~\citep{hunyuan3d}.
To handle the temporal dependencies inherent in motion (e.g., gait phases), we precede the DiT with a \textit{Context Refinement Transformer}. This lightweight module processes $\mathbf{C}_{\text{vis}}$ with alternating temporal-attention (modeling frame-to-frame coherence) and global-attention (aggregating sequence-level context), producing a stabilized visual signal $\mathbf{C}^*$ for the flow matching process.

%% file: experiments_and_discussion.tex
\section{Experiments}
We validate the effectiveness of \MethodMotion through comprehensive quantitative and qualitative evaluations. Our experiments are designed to investigate three key hypotheses:
(1) That a single unified model can reconstruct motion with fidelity comparable to specialist models trained on specific topologies (Sec.~\ref{sec:exp_recon}).
(2) That the learned manifold generalizes effectively to unseen skeletal structures, i.e., zero-shot topology (Sec.~\ref{sec:exp_gen}).
(3) That the generative extraction remains robust under sparse or noisy visual evidence (Sec.~\ref{sec:exp_ablation}).

\subsection{Experimental Setup}
\paragraph{Evaluation Metrics}
We employ four complementary metrics to assess reconstruction fidelity and kinematic validity.
\begin{itemize}
    \item \textbf{Mean Per Joint Position Error (MPJPE)}~\citep{videopose3d}: Measures the Euclidean distance between predicted and ground-truth joint positions after Forward Kinematics (FK), averaged over all joints and frames.
    \item \textbf{Mean Per Joint Velocity Error (MPJVE)}~\citep{videopose3d}: Assesses temporal consistency and stability by measuring the discrepancy in joint velocities.
    \item \textbf{Chamfer Distance (CD)}~\citep{xu2020rignet}: Since topology varies, standard joint-to-joint metrics are insufficient for cross-topology comparisons. We use two-sided CD to measure the geometric similarity between the generated and ground-truth skeletal point clouds.
    \item \textbf{Geodesic Distance (GD)}~\citep{he2022nemf}: To explicitly evaluate rotational accuracy independent of bone lengths, we compute the minimal geodesic distance on the $\mathrm{SO}(3)$ manifold. This metric is crucial for characterizing the quality of the learned local rotation priors.
\end{itemize}
Please refer to the Supplementary C.4 for formal definitions.
\subsection{Results and Comparison}

\subsubsection{Motion Reconstruction} \label{sec:exp_recon}
We first evaluate the representational capacity of our CVAE. \textit{To our knowledge, there are no open-source methods specifically designed for topology-agnostic motion reconstruction, as most existing approaches target humans or quadruped animals with fixed skeletal templates.} Consequently, to provide a rigorous evaluation, we benchmark against state-of-the-art specialist models on their respective domains.

\paragraph{Baselines}
We compare against MotionMillion~\citep{fan2025zerozeroshotmotiongeneration}, MoMask~\citep{guo2024momask}, and MoMADiff~\citep{zhang2025towards} on the human-centric HumanML3D dataset~\citep{humanml3d}, and AniMo~\citep{wang2025animo} on the quadruped-specific AniMo4D dataset~\citep{wang2025animo}. We also test on the topology-agnostic Truebones Zoo~\citep{truebones} and our held-out \MethodData.
\paragraph{Analysis.}
As reported in Tab.~\ref{tab:result_cvae}, our method consistently outperforms baselines in both MPJPE and MPJVE. Remarkably, \textsc{TopoCap} achieves superior reconstruction accuracy even on datasets with fixed templates (HumanML3D and AniMo4D), surpassing specialist models trained on those topologies. On the highly diverse Truebones Zoo and \MethodData, where template-based methods are inapplicable, our approach maintains high fidelity. This confirms that our disentanglement of \textit{structure} (skeleton) and \textit{dynamics} (motion) allows the model to learn a shared, high-quality motion manifold without suffering from negative transfer or capacity dilution.

\begin{table}[h!]
    \centering
    \caption{\textbf{Motion reconstruction fidelity.} Comparison against specialist models. Missing entries ($/$) indicate unsupported topologies. \MethodMotion~achieves state-of-the-art performance on human and quadruped benchmarks while uniquely handling the structural diversity of \MethodData.}
    \begin{tabular}{c|rr|rr}
    \hline
        & \multicolumn{2}{c|}{\textbf{HumanML3D}} & \multicolumn{2}{c}{\textbf{AniMo4D}} \\
    \cline{2-5}
        & MPJPE$\downarrow$ & MPJVE$\downarrow$ & MPJPE$\downarrow$ & MPJVE$\downarrow$ \\
    \hline
        MotionMillion & 10.48 & 8.64  & /    & /    \\
        MoMask        & 14.34 & 11.28 & /    & /    \\
        MoMADiff      & 9.32  & 5.17  & /    & /    \\
        AniMo         & /     & /     & 12.11& 9.82 \\
        Ours          & \final{\textbf{5.44}} & \final{\textbf{1.32}} & \final{\textbf{5.78}} & \final{\textbf{1.07}} \\
    \hline
        & \multicolumn{2}{c|}{\textbf{Truebones Zoo}} & \multicolumn{2}{c}{\textbf{\MethodData}} \\
    \cline{2-5}
        & MPJPE$\downarrow$ & MPJVE$\downarrow$ & MPJPE$\downarrow$ & MPJVE$\downarrow$ \\
    \hline
        Ours          & \final{\textbf{6.88}} & \final{\textbf{3.74}} & \final{\textbf{20.53}} & \final{\textbf{8.78}} \\
    \hline
    \end{tabular}

    \label{tab:result_cvae}
\end{table}

\subsubsection{Generative Motion Extraction} \label{sec:exp_gen}
We next evaluate the full video-to-animation pipeline. This task requires the model to synthesize plausible 3D motion \final{from 2D video} while strictly adhering to the kinematic constraints of an arbitrary target skeleton.

\paragraph{Baselines}
We compare against two categories of approaches:
(1) \textit{Puppeteer}~\citep{song2025puppeteer}, an optimization-based method relying on optical flow.
(2) \textit{GenZoo}~\citep{genzoo}, a learning-based method targeting quadrupeds.
We evaluate on Truebones Zoo and \MethodData. Crucially, we split the test data into \textit{Seen} (topologies present during training) and \textit{Unseen} (novel topologies never seen by the network) to rigorously assess generalization. \final{We further use 5\% of the total data for validation, evenly split between seen and unseen topologies.}

\paragraph{Analysis}
Quantitative results are presented in Tab.~\ref{tab:result_diffusion}. Optimization-based methods like Puppeteer struggle with depth ambiguity and local minima, leading to high error rates. GenZoo performs adequately on quadrupeds but fails to generalize to the long-tail diverse creatures in our dataset.
In contrast, \MethodMotion~ achieves state-of-the-art performance. Most importantly, the performance gap between \textit{Seen} and \textit{Unseen} topologies is marginal. This serves as strong empirical evidence that our model has learned a truly universal motion prior rather than simply memorizing specific skeletal configurations. Qualitative results in Fig.~\ref{fig:qual_comp} further illustrate that our method produces temporally coherent, physically plausible motions where baselines often exhibit jitter or catastrophic structural failure.

\begin{table}[t]
\centering
\caption{\textbf{Quantitative comparison of video-to-motion extraction.} Performance on \textbf{Seen} (training topologies) and \textbf{Unseen} (zero-shot) splits. \MethodMotion~outperforms optimization-based and generative baselines, with a marginal Seen–Unseen gap, confirming that \MethodMotion~learns generalizable motion priors rather than memorizing specific skeletons.}
\small
\setlength{\tabcolsep}{4pt}
\begin{tabular}{c|c|rrrr}
\hline
Dataset & Method & MPJPE$\downarrow$ & MPJVE$\downarrow$ & CD$\downarrow$ & GD$\downarrow$ \\
\hline
\multirow{4}{*}{\textbf{Truebones Zoo}}
& Puppeteer   & 227.21 & 87.85 & 0.095 & 0.622 \\
& GenZoo      & /      & /     & 0.114 & /     \\
& Ours (Seen)  & \final{\textbf{52.05}}  & \final{12.05} & \final{\textbf{0.049}} & \final{\textbf{0.289}} \\
& Ours (Unseen)& \final{62.54}  & \final{\textbf{8.67}}  & \final{0.057} & \final{0.529} \\
\hline
\multirow{4}{*}{\textbf{\MethodData}}
& Puppeteer   & \final{342.52} & \final{108.628} & \final{0.117} & \final{0.694} \\
& GenZoo      & /      & /     & \final{0.145} & /     \\
& Ours (Seen)  & \final{\textbf{179.61}} & \final{\textbf{47.99}} & \final{\textbf{0.092}} & \final{0.487} \\
& Ours (Unseen)& \final{192.04} & \final{51.76} & \final{0.102} & \final{\textbf{0.463}} \\
\hline
\end{tabular}

\label{tab:result_diffusion}
\end{table}

\subsection{Ablation Study} \label{sec:exp_ablation}
We conduct ablation studies to validate our architectural choices (Tab.~\ref{tab:ablation_cvae}) and the robustness of our diffusion sampling (Tab.~\ref{tab:ablation_diffusion}).
\paragraph{Impact of Reconstruction Components} \label{ablation:cvae}
We analyze three key CVAE components: (1) \textit{Physics-Aware Output:} Replacing global decoding with local regression causes the largest degradation (Tab.~\ref{tab:ablation_cvae}), highlighting error accumulation along kinematic chains, which our formulation effectively mitigates. (2) \textit{Temporal Attention:} Removing temporal modeling sharply increases MPJVE, indicating the importance of frame dependencies for smooth motion synthesis. (3) \textit{Velocity Loss:} Ablating $\mathcal{L}_{vel}$ introduces high-frequency jitter, confirming its role as a regularization term in stabilizing dynamics.

\begin{table}[th]
    \centering
    % \vspace{-1mm}
    \caption{\textbf{Ablation of Manifold Learning components.} We validate the necessity of our architectural choices. \textit{Global-First Decoding} (removing ``w/o absolute decoding'') is critical for reducing MPJPE, as purely local predictions drift. Removing temporal attention degrades smoothness (MPJVE), confirming its role in modeling dynamics.}
    % \vspace{-2mm}
    \begin{tabular}{c|rr|rr}
    \hline
        & \multicolumn{2}{c|}{\textbf{Truebones Zoo}} & \multicolumn{2}{c}{\textbf{\MethodData}}\\
    \cline{2-5}
        & MPJPE$\downarrow$ & MPJVE$\downarrow$ & MPJPE$\downarrow$ & MPJVE$\downarrow$ \\
    \hline
        w/o absolute decoding & \final{10.32} & \final{8.45} & \final{33.74} & \final{14.62} \\
        w/o temporal attention & \final{8.97} & \final{5.32} & \final{23.74} & \final{10.73} \\
        w/o velocity loss term & \final{7.18} & \final{4.11} & \final{22.03} & \final{9.14} \\
        Ours          & \final{\textbf{6.88}} & \final{\textbf{3.74}} & \final{\textbf{20.53}} & \final{\textbf{8.78}} \\
    \hline
    \end{tabular}
    \label{tab:ablation_cvae}
\end{table}

\begin{table}[th]
    \centering
    \caption{\textbf{Robustness to Sparse Visual Evidence.}
    We downsample the input video (stride $X$), observing only 1 frame every $X$ frames. Marginal performance degradation even at $8\times$ demonstrates that our generative prior plausibly infills missing dynamics.}
    \begin{tabular}{c|cccc}
    \hline
    \multicolumn{5}{c}{\textbf{Truebones Zoo}} \\
    \hline
        & MPJPE$\downarrow$ & MPJVE$\downarrow$ & CD$\downarrow$ & GD$\downarrow$ \\
    \hline
        Stride 1 (baseline)      & \final{\textbf{52.05}}  & \final{\textbf{12.05}} & \final{\textbf{0.049}} & \final{\textbf{0.289}} \\
        Stride 4  & \final{54.71} & \final{13.80} & \final{0.052} & \final{0.292} \\
        Stride 8 & \final{55.90} & \final{15.32} & \final{0.054} & \final{0.331} \\
    \hline
    \hline
    \multicolumn{5}{c}{\textbf{\MethodData}} \\
    \hline
        & MPJPE$\downarrow$ & MPJVE$\downarrow$ & CD$\downarrow$ & GD$\downarrow$ \\
    \hline
        Stride 1 (baseline)  & \final{\textbf{179.61}} & \final{\textbf{47.99}} & \final{\textbf{0.092}} & \final{\textbf{0.487}} \\
        Stride 4  & \final{187.34} & \final{50.06} & \final{0.101} & \final{0.496} \\
        Stride 8 & \final{193.54} & \final{52.62} & \final{0.117} & \final{0.510} \\
    \hline
    \end{tabular}
    % \vspace{-4mm}
    \label{tab:ablation_diffusion}
\end{table}

\paragraph{Robustness to Sparse Visual Conditioning}
A robust motion capture system should handle temporal occlusion and low-framerate inputs. We evaluate our flow model's ability to act as a ``motion in-betweening" engine by providing visual conditioning only at sparse keyframes, forcing the model to infer intermediate dynamics.
As shown in Tab.~\ref{tab:ablation_diffusion}, performance degrades only marginally even when visual information is reduced by 8$\times$. This indicates that the generative model effectively leverages the learned motion prior to fill in missing temporal information, relying on the latent manifold's continuity rather than purely on per-frame visual cues.

\begin{table}[t]
\final{
    \centering
    \caption{\final{\textbf{Effect of Latent Motion Representation.}
    Comparison with direct prediction under the AnyTop protocol. ``w/ latent'' uses our CVAE representation, while ``w/o latent'' predicts motion tokens directly. Removing rest-pose conditioning causes severe degradation, especially on unseen topologies.}}
    \small
    % \vspace{-2mm}
    \begin{tabular}{c|c|cccc}
    \hline
    \multicolumn{6}{c}{\textbf{Truebones Zoo}} \\
    \hline
    Split & Method & MPJPE$\downarrow$ & MPJVE$\downarrow$ & CD$\downarrow$ & GD$\downarrow$ \\
    \hline
    \multirow{3}{*}{Seen}
    & Ours & \textbf{52.05} & 12.05 & \textbf{0.049} & \textbf{0.289} \\
    & Baseline (w/ latent) & 57.82 & \textbf{11.31} & 0.053 & 0.327 \\
    & Baseline (w/o latent) & 115.09 & 20.67 & 0.088 & 0.771 \\
    \hline
    \multirow{3}{*}{Unseen}
    & Ours & \textbf{62.54} & \textbf{8.67} & \textbf{0.057} & \textbf{0.529} \\
    & Baseline (w/ latent) & 66.32 & 10.48 & \textbf{0.057} & 0.629 \\
    & Baseline (w/o latent) & 210.58 & 18.08 & 0.151 & 0.871 \\
    \hline
    \hline
    \multicolumn{6}{c}{\textbf{\MethodData}} \\
    \hline
    Split & Method & MPJPE$\downarrow$ & MPJVE$\downarrow$ & CD$\downarrow$ & GD$\downarrow$ \\
    \hline
    \multirow{3}{*}{Seen}
    & Ours & \textbf{179.61} & 47.99 & 0.092 & \textbf{0.487} \\
    & Baseline (w/ latent) & 187.38 & \textbf{46.35} & \textbf{0.087} & 0.570 \\
    & Baseline (w/o latent) & 349.74 & 62.91 & 0.150 & 0.684 \\
    \hline
    \multirow{3}{*}{Unseen}
    & Ours & \textbf{192.04} & 51.76 & 0.102 & \textbf{0.463} \\
    & Baseline (w/ latent) & 198.77 & \textbf{49.92} & \textbf{0.096} & 0.514 \\
    & Baseline (w/o latent) & 357.67 & 69.76 & 0.171 & 0.819 \\
    \hline
    \end{tabular}
    % \vspace{-4mm}
    \label{tab:latent_ablation}
}
\end{table}

\final{
\paragraph{Impact of Latent Representation on Motion Generation.}
To validate the proposed topology-aware latent space (Sec.~\ref{sec:cvae}), we compare against two joint-space baselines following AnyTop~\citep{anytop}: one with rest-pose conditioning (Sec.~\ref{sec:conditioning}) and one without. As shown in Tab.~\ref{tab:latent_ablation}, our latent formulation consistently achieves lower MPJPE and GD on both seen and unseen datasets, indicating improved pose accuracy and global structure preservation. Removing rest-pose conditioning results in significant degradation across all metrics. These findings suggest that the proposed latent representation effectively disentangles motion from skeletal structure, leading to stronger generalization.
}

\section{Applications}

\subsection{Zero-Shot Motion Retargeting}
A compelling emergent property of \MethodMotion is its ability to perform zero-shot motion retargeting. Although not trained with a retargeting loss, our disentangled representation, where $z$ encodes dynamics and $\mathcal{S}$ encodes structure, allows swapping $\mathcal{S}_{\text{target}}$ while preserving $z_{\text{source}}$, as shown in Fig.~\ref{fig:retarget}. We transfer motion from a source video to a target skeleton with a radically different topology (e.g., quadruped to avian), preserving high-level semantics (e.g., rhythm) while adapting low-level kinematics, suggesting the latent space captures abstract locomotion beyond simple joint correlations.
\begin{figure}[t]
    \includegraphics[width=0.95\columnwidth]{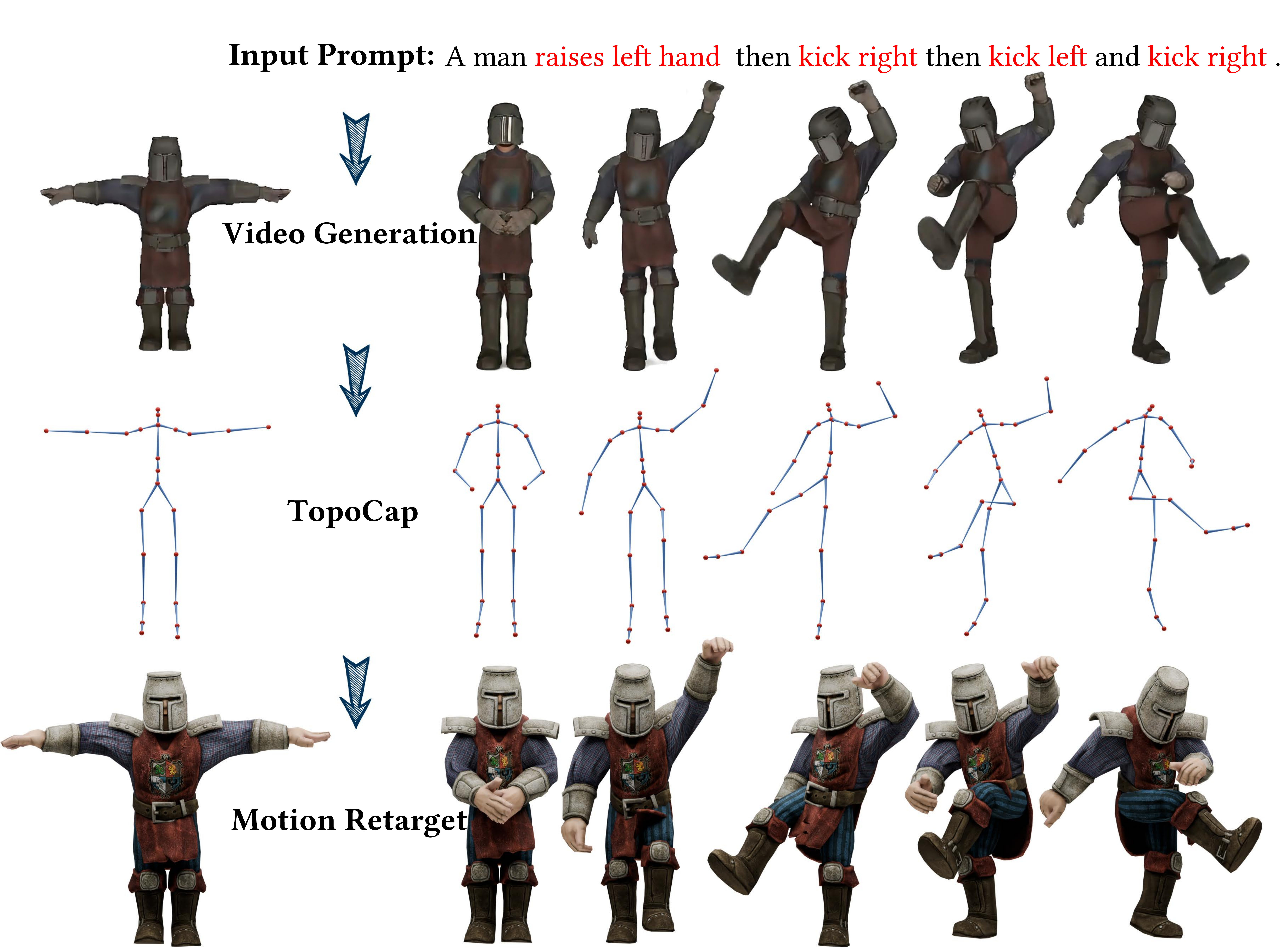}
    \caption{\textbf{Application: Scalable 4D Generation via Video Models.} By chaining a Text-to-Video model (Wan2.1~\citep{wan2025}) with \MethodMotion, we enable text-to-animation for arbitrary characters, turning video generators into scalable sources of 3D motion data.}
    
    \label{fig:wild}
\end{figure}

\begin{figure}[t]
    \centering
    \includegraphics[width=0.95\columnwidth]{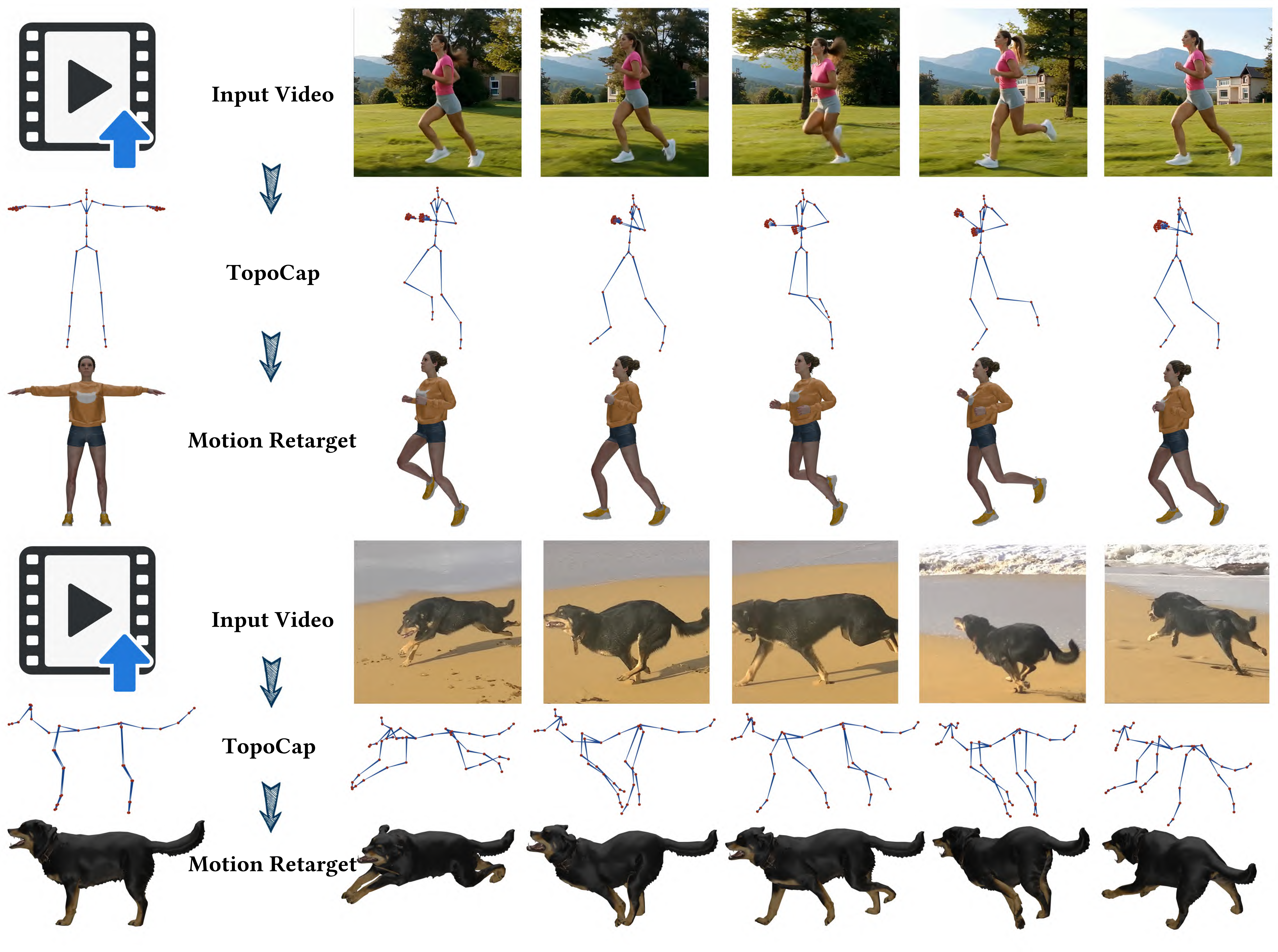}
    \caption{\final{\textbf{Real-World Mocap.} Given a rigged 3D asset, \MethodMotion~ directly extracts 3D motion from real-world videos.}}
    \label{fig:realworld}
\end{figure}

\begin{figure}[t]
    \centering
    % \vspace{-3mm}
    \includegraphics[width=0.95\columnwidth]{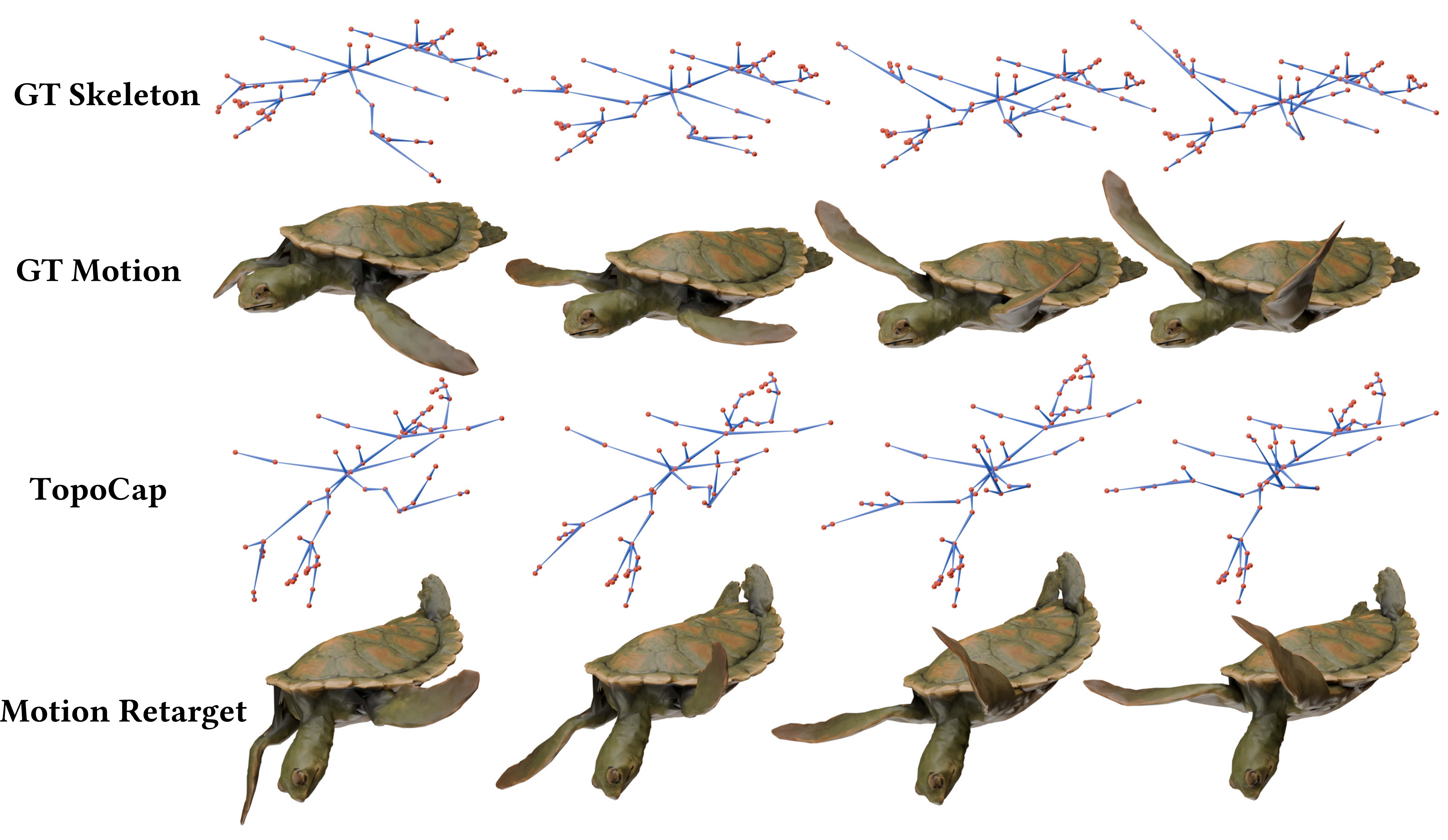}
    \caption{\final{\textbf{Failure case.}} \final{When the target topology is highly uncommon, the extracted motion may exhibit significant deviations.}}
    \label{fig:fail}
\end{figure}

\subsection{Scalable Data Generation}
The scarcity of diverse 3D motion data is a primary hindrance in data-driven character animation. Our framework serves as a scalable engine for \final{converting synthetic or monocular videos} into rigged 3D motion data.
By combining a video generation model (e.g., Wan2.1~\citep{wan2025}) with \MethodMotion, we establish a text-to-animation pipeline (Fig.~\ref{fig:wild}) that generates videos from text and extracts high-quality 3D motion for arbitrary rigs. This eliminates the need for MoCap hardware or manual keyframing, enabling large-scale, diverse motion datasets (including non-humanoid long-tail cases) for training motion foundation models. \final{\MethodMotion~ also generalizes to real-world videos (Fig.~\ref{fig:realworld}), demonstrating strong robustness.}
\section{Conclusion}
We presented \MethodMotion, \final{a unified framework that learns topology-agnostic motion priors for video-driven 3D animation}. By treating skeletal structure as conditions rather than constraints, we learn a universal motion manifold that animates arbitrary characters without test-time optimization or template-specific training, enabled by \MethodData, the curated structurally diverse motion dataset. Experiments show strong generalization to unseen topologies and support applications like zero-shot retargeting and motion data generation, advancing 3D animation toward universal motion understanding.

\noindent\textit{Limitations and future work.} \final{As shown in Fig.~\ref{fig:fail}, performance degrades on highly uncommon topologies, and the method is sensitive to input video quality and domain gaps in real-world footage. It also requires a predefined skeleton and operates in camera space without modeling global trajectories or physical constraints (e.g., foot contact).} Future work will explore end-to-end motion generation under diverse conditions, with improved physical modeling and reduced reliance on dense visual inputs.
\begin{acks}
This work was supported by Fundamental and Interdisciplinary Disciplines Breakthrough Plan of the Ministry of Education of China (No.JYB2025XDXM101), the National Natural Science Foundation of China (No.62220106003), and the Research Grant of Tsinghua-Tencent Joint Laboratory for Internet Innovation Technology. Yan-Pei Cao was supported by Beijing Major Science and Technology Project under Contract (No.Z251100007125016), and the International (Hong Kong, Macao, and Taiwan) Collaborative R\&D Project. We also thank Ming-Yuan Zhang for his insightful advice.

\end{acks}

% Bibliography
\bibliographystyle{ACM-Reference-Format}
\bibliography{bibliography}

% Appendix

% \input{supp/supp}

% \vspace{-5mm}
\begin{figure*}
    \centering
    \includegraphics[width=\textwidth]{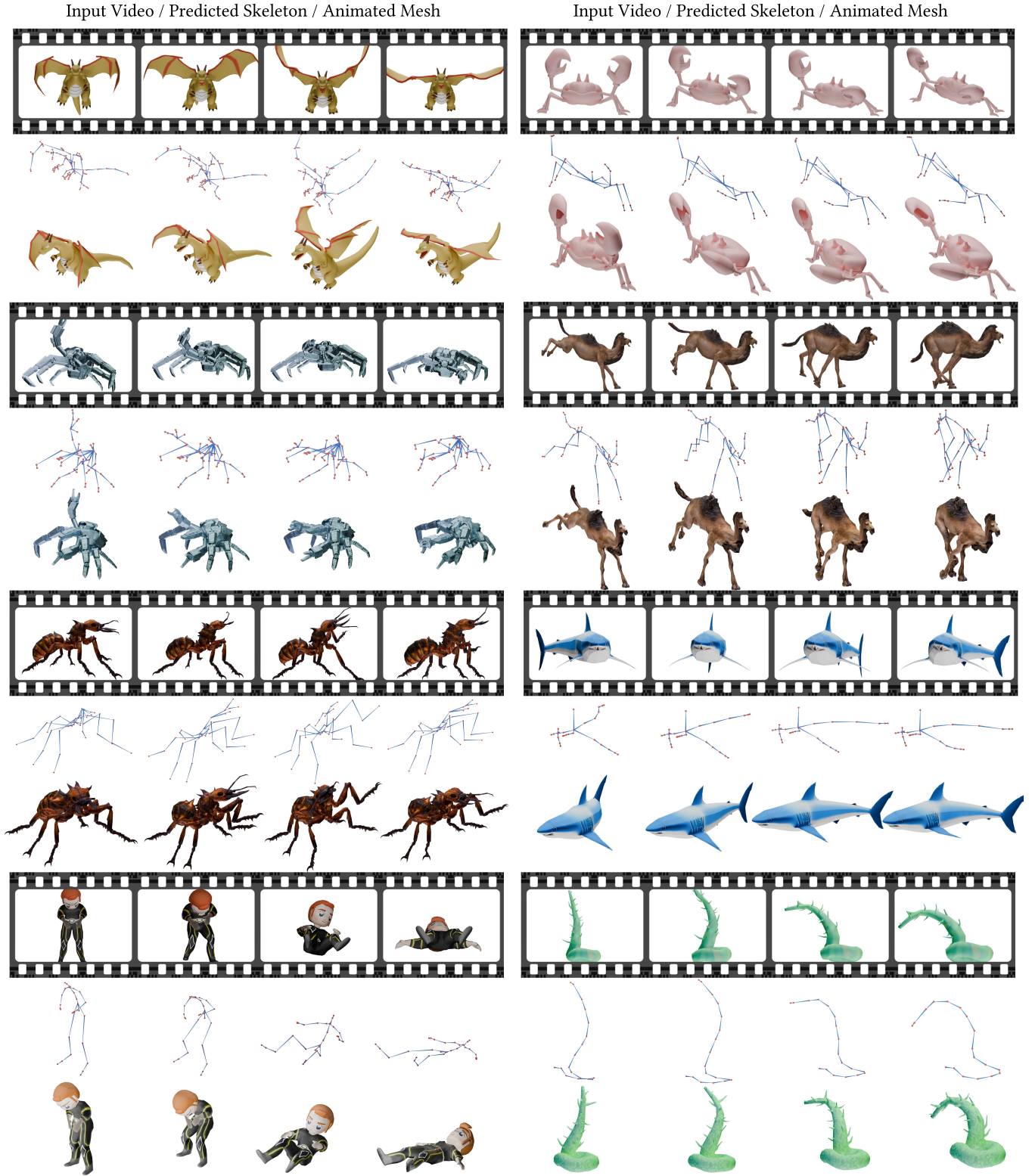}
    \caption{\textbf{Zero-Shot Motion Extraction.} Given monocular videos, \MethodMotion~accurately predicts the articulation for diverse creatures. Note the structural variety: from multi-legged insects to finned aquatic life, the model respects the distinct kinematic constraints of each rig.}
    \label{fig:video_result}
\end{figure*}

\begin{center}
\begin{figure*}
    \captionsetup{skip=1pt}
    \includegraphics[width=0.85\textwidth]{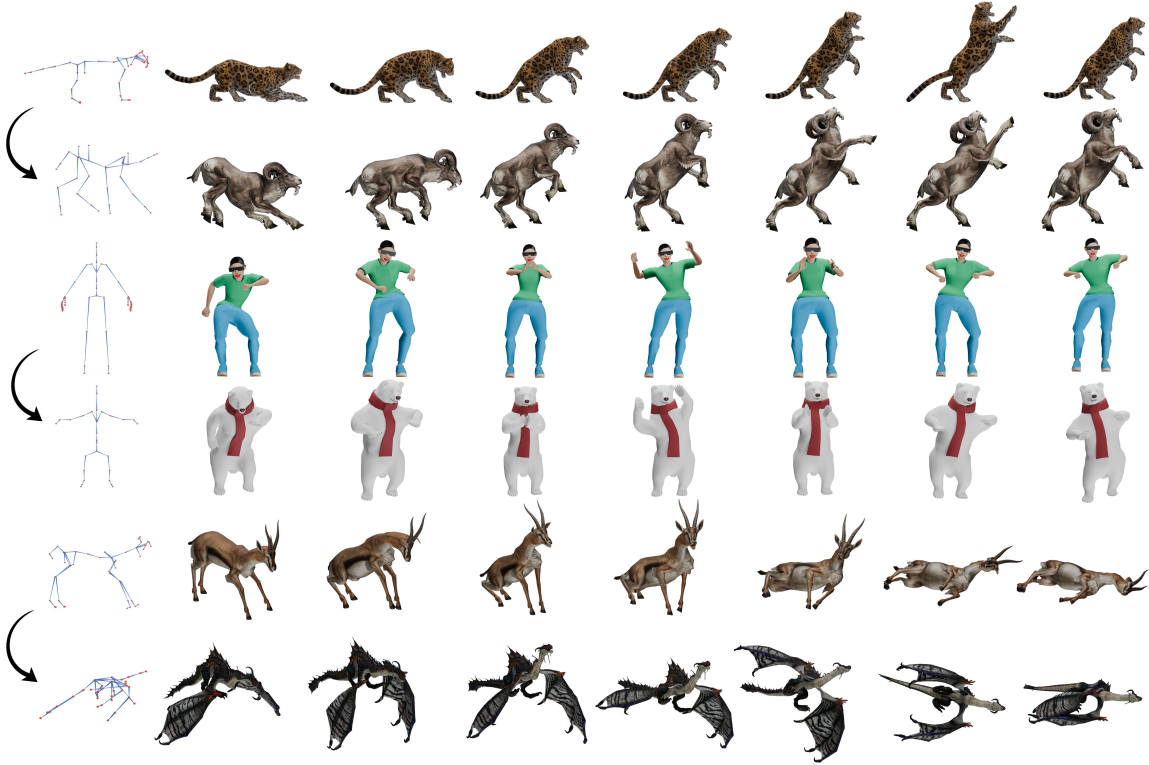}
    \caption{\textbf{Cross-Topology Motion Retargeting.} By swapping the target rig condition $\mathcal{S}$, we can transfer motion from a source character (Top) to a radically different target (Bottom). The model preserves high-level semantics (gait, energy, phase) while adapting low-level kinematics to the new body plan (e.g., adapting a quadrupedal falling to a flying dragon).}
    \label{fig:retarget}
\end{figure*}

\begin{figure*}
    \captionsetup{skip=1pt}
    \includegraphics[width=0.95\linewidth]{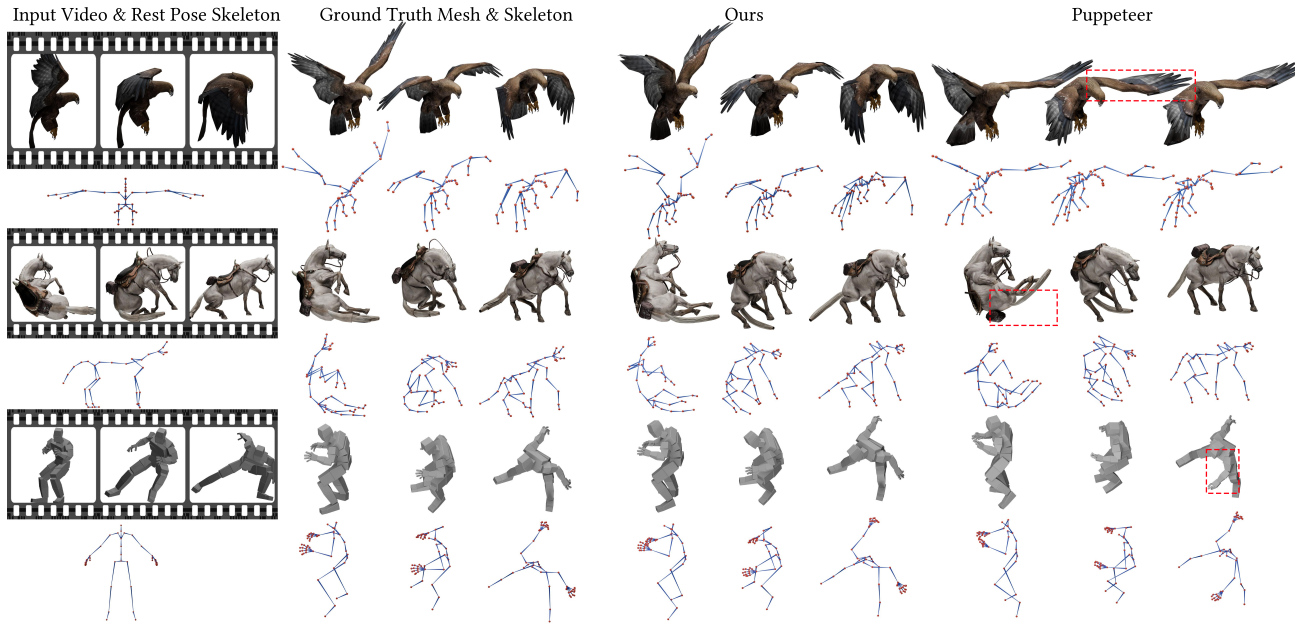}
    \caption{\textbf{Visual Benchmark vs. Optimization (Puppeteer).} We visualize reconstructions from a novel top-down viewpoint to reveal 3D quality (inputs are front-view). Our method (Center) leverages the learned manifold to resolve depth ambiguities, producing structurally valid poses. In contrast, Puppeteer (Right) relies on 2D projection constraints, leading to severe depth artifacts highlighted in red: note the \textit{unnatural wing twisting} (Top), \textit{limb collapse} (Middle), and \textit{joint dislocation} (Bottom).}
    \label{fig:qual_comp}
\end{figure*}
\end{center}

%% file: bibliography.bib
@article{smpl,
author = {Loper, Matthew and Mahmood, Naureen and Romero, Javier and Pons-Moll, Gerard and Black, Michael J.},
title = {SMPL: a skinned multi-person linear model},
year = {2015},
issue_date = {November 2015},
publisher = {Association for Computing Machinery},
address = {New York, NY, USA},
volume = {34},
number = {6},
issn = {0730-0301},
url = {https://doi.org/10.1145/2816795.2818013},
doi = {10.1145/2816795.2818013},
journal = {ACM Trans. Graph.},
month = nov,
articleno = {248},
numpages = {16}
}

@misc{mhr,
      title={MHR: Momentum Human Rig}, 
      author={Aaron Ferguson and Ahmed A. A. Osman and Berta Bescos and Carsten Stoll and Chris Twigg and Christoph Lassner and David Otte and Eric Vignola and Fabian Prada and Federica Bogo and Igor Santesteban and Javier Romero and Jenna Zarate and Jeongseok Lee and Jinhyung Park and Jinlong Yang and John Doublestein and Kishore Venkateshan and Kris Kitani and Ladislav Kavan and Marco Dal Farra and Matthew Hu and Matthew Cioffi and Michael Fabris and Michael Ranieri and Mohammad Modarres and Petr Kadlecek and Rawal Khirodkar and Rinat Abdrashitov and Romain Prévost and Roman Rajbhandari and Ronald Mallet and Russell Pearsall and Sandy Kao and Sanjeev Kumar and Scott Parrish and Shoou-I Yu and Shunsuke Saito and Takaaki Shiratori and Te-Li Wang and Tony Tung and Yichen Xu and Yuan Dong and Yuhua Chen and Yuanlu Xu and Yuting Ye and Zhongshi Jiang},
      year={2025},
      eprint={2511.15586},
      archivePrefix={arXiv},
      primaryClass={cs.GR},
      url={https://arxiv.org/abs/2511.15586}, 
}

@article{amp,
author = {Peng, Xue Bin and Ma, Ze and Abbeel, Pieter and Levine, Sergey and Kanazawa, Angjoo},
title = {AMP: adversarial motion priors for stylized physics-based character control},
year = {2021},
issue_date = {August 2021},
publisher = {Association for Computing Machinery},
address = {New York, NY, USA},
volume = {40},
number = {4},
issn = {0730-0301},
url = {https://doi.org/10.1145/3450626.3459670},
doi = {10.1145/3450626.3459670},
journal = {ACM Trans. Graph.},
month = jul,
articleno = {144},
numpages = {20}
}

@inproceedings{ppmotion,
author = {Zhao, Sihan and Wang, Zixuan and Luan, Tianyu and Jia, Jia and Zhu, Wentao and Luo, Jiebo and Yuan, Junsong and Xi, Nan},
title = {PP-Motion: Physical-Perceptual Fidelity Evaluation for Human Motion Generation},
year = {2025},
isbn = {9798400720352},
publisher = {Association for Computing Machinery},
address = {New York, NY, USA},
url = {https://doi.org/10.1145/3746027.3754940},
doi = {10.1145/3746027.3754940},
booktitle = {Proceedings of the 33rd ACM International Conference on Multimedia},
pages = {6840–6849},
numpages = {10},
location = {Dublin, Ireland},
series = {MM '25}
}

@inproceedings{skillmimicv2,
author = {Yu, Runyi and Wang, Yinhuai and Zhao, Qihan and Tsui, Hok Wai and Wang, Jingbo and Tan, Ping and Chen, Qifeng},
title = {SkillMimic-V2: Learning Robust and Generalizable Interaction Skills from Sparse and Noisy Demonstrations},
year = {2025},
isbn = {9798400715402},
publisher = {Association for Computing Machinery},
address = {New York, NY, USA},
url = {https://doi.org/10.1145/3721238.3730640},
doi = {10.1145/3721238.3730640},
booktitle = {Proceedings of the Special Interest Group on Computer Graphics and Interactive Techniques Conference Conference Papers},
articleno = {115},
numpages = {11},
location = {
},
series = {SIGGRAPH Conference Papers '25}
}

@article{deepmimic,
author = {Peng, Xue Bin and Abbeel, Pieter and Levine, Sergey and van de Panne, Michiel},
title = {DeepMimic: example-guided deep reinforcement learning of physics-based character skills},
year = {2018},
issue_date = {August 2018},
publisher = {Association for Computing Machinery},
address = {New York, NY, USA},
volume = {37},
number = {4},
issn = {0730-0301},
url = {https://doi.org/10.1145/3197517.3201311},
doi = {10.1145/3197517.3201311},
month = jul,
articleno = {143},
numpages = {14}
}

@article{dong2020,
author = {Dong, Junting and Shuai, Qing and Sun, Jingxiang and Zhang, Yuanqing and Bao, Hujun and Zhou, Xiaowei},
title = {iMoCap: Motion Capture from Internet Videos},
year = {2022},
issue_date = {May 2022},
publisher = {Kluwer Academic Publishers},
address = {USA},
volume = {130},
number = {5},
issn = {0920-5691},
url = {https://doi.org/10.1007/s11263-022-01596-7},
doi = {10.1007/s11263-022-01596-7},
journal = {Int. J. Comput. Vision},
month = may,
pages = {1165–1180},
numpages = {16}
}

@inproceedings{kocabas2020,
  title={Vibe: Video inference for human body pose and shape estimation},
  author={Kocabas, Muhammed and Athanasiou, Nikos and Black, Michael J},
  booktitle={Proceedings of the IEEE/CVF conference on computer vision and pattern recognition},
  pages={5253--5263},
  year={2020}
}

@inproceedings{AnimalAvatars2024,
author = {Sabathier, Remy and Mitra, Niloy J. and Novotny, David},
title = {Animal Avatars: Reconstructing Animatable 3D Animals from Casual Videos},
year = {2024},
booktitle = {Computer Vision – ECCV 2024: 18th European Conference, Milan, Italy, September 29–October 4, 2024, Proceedings, Part LXXIX},
pages = {270–287},
doi = {10.1007/978-3-031-72986-7_16}
}

@inproceedings{rempe2021,
    author={Rempe, Davis and Birdal, Tolga and Hertzmann, Aaron and Yang, Jimei and Sridhar, Srinath and Guibas, Leonidas J.},
    title={HuMoR: 3D Human Motion Model for Robust Pose Estimation},
    booktitle={International Conference on Computer Vision (ICCV)},
    year={2021}
}

@inproceedings{wang2025animo,
  title={AniMo: Species-Aware Model for Text-Driven Animal Motion Generation},
  author={Wang, Xuan and Ruan, Kai and Zhang, Xing and Wang, Gaoang},
  booktitle={Proceedings of the Computer Vision and Pattern Recognition Conference},
  pages={1929--1939},
  year={2025}
}

@inproceedings{yang2023omnimotiongpt,
  author       = {Zhangsihao Yang and
                  Mingyuan Zhou and
                  Mengyi Shan and
                  Bingbing Wen and
                  Ziwei Xuan and
                  Mitch Hill and
                  Junjie Bai and
                  Guo{-}Jun Qi and
                  Yalin Wang},
  title        = {OmniMotionGPT: Animal Motion Generation with Limited Data},
  booktitle    = {{IEEE/CVF} Conference on Computer Vision and Pattern Recognition,
                  {CVPR} 2024, Seattle, WA, USA, June 16-22, 2024},
  pages        = {1249--1259},
  publisher    = {{IEEE}},
  year         = {2024},
  url          = {https://doi.org/10.1109/CVPR52733.2024.00125},
  doi          = {10.1109/CVPR52733.2024.00125},
  bibsource    = {dblp computer science bibliography, https://dblp.org}
}

@misc{deb2025articulate3dzeroshottextdriven3d,
      title={Articulate3D: Zero-Shot Text-Driven 3D Object Posing}, 
      author={Oishi Deb and Anjun Hu and Ashkan Khakzar and Philip Torr and Christian Rupprecht},
      year={2025},
      eprint={2508.19244},
      archivePrefix={arXiv},
      primaryClass={cs.CV},
      url={https://arxiv.org/abs/2508.19244}, 
}

@misc{smoogpt,
      title={SMooGPT: Stylized Motion Generation using Large Language Models}, 
      author={Lei Zhong and Yi Yang and Changjian Li},
      year={2025},
      eprint={2509.04058},
      archivePrefix={arXiv},
      primaryClass={cs.GR},
      url={https://arxiv.org/abs/2509.04058}, 
}

@misc{xmogen,
      title={X-MoGen: Unified Motion Generation across Humans and Animals}, 
      author={Xuan Wang and Kai Ruan and Liyang Qian and Zhizhi Guo and Chang Su and Gaoang Wang},
      year={2025},
      eprint={2508.05162},
      archivePrefix={arXiv},
      primaryClass={cs.CV},
      url={https://arxiv.org/abs/2508.05162}, 
}

@article{song2025puppeteer,
  title={Puppeteer: Rig and Animate Your 3D Models},
  author={Chaoyue Song and Xiu Li and Fan Yang and Zhongcong Xu and Jiacheng Wei and Fayao Liu and Jiashi Feng and Guosheng Lin and Jianfeng Zhang},
  journal={arXiv preprint arXiv:2508.10898},
  year={2025}
}

@INPROCEEDINGS {atom,
author = { Han, Haonan and Wu, Xiangzuo and Liao, Huan and Xu, Zunnan and Hu, Zhongyuan and Li, Ronghui and Zhang, Yachao and Li, Xiu },
booktitle = { 2025 IEEE/CVF Conference on Computer Vision and Pattern Recognition (CVPR) },
title = {{ AToM: Aligning Text-to-Motion Model at Event-Level with GPT-4Vision Reward }},
year = {2025},
volume = {},
ISSN = {},
pages = {22746-22755},
doi = {10.1109/CVPR52734.2025.02118},
url = {https://doi.ieeecomputersociety.org/10.1109/CVPR52734.2025.02118},
publisher = {IEEE Computer Society},
address = {Los Alamitos, CA, USA},
month =Jun}

@inproceedings{liao2025shape,
  title={Shape my moves: Text-driven shape-aware synthesis of human motions},
  author={Liao, Ting-Hsuan and Zhou, Yi and Shen, Yu and Huang, Chun-Hao Paul and Mitra, Saayan and Huang, Jia-Bin and Bhattacharya, Uttaran},
  booktitle={Proceedings of the Computer Vision and Pattern Recognition Conference},
  pages={1917--1928},
  year={2025}
}

@inproceedings{raab2024single,
title={Single Motion Diffusion},
author={Raab, Sigal and Leibovitch, Inbal and Tevet, Guy and Arar, Moab and Bermano, Amit H and Cohen-Or, Daniel},
booktitle={The Twelfth International Conference on Learning Representations (ICLR)},             
url={https://openreview.net/pdf?id=DrhZneqz4n},
year={2024}
}

@article{ganimator,
author = {Li, Peizhuo and Aberman, Kfir and Zhang, Zihan and Hanocka, Rana and Sorkine-Hornung, Olga},
title = {GANimator: neural motion synthesis from a single sequence},
year = {2022},
issue_date = {July 2022},
publisher = {Association for Computing Machinery},
address = {New York, NY, USA},
volume = {41},
number = {4},
issn = {0730-0301},
url = {https://doi.org/10.1145/3528223.3530157},
doi = {10.1145/3528223.3530157},
journal = {ACM Trans. Graph.},
month = jul,
articleno = {138},
numpages = {12}
}

@inproceedings{animax,
author = {Huang, Zehuan and Feng, Haoran and Sun, Yang-Tian and Guo, Yuan-Chen and Cao, Yan-Pei and Sheng, Lu},
title = {AnimaX: Animating the Inanimate in 3D with Joint Video-Pose Diffusion Models},
year = {2025},
isbn = {9798400721373},
publisher = {Association for Computing Machinery},
address = {New York, NY, USA},
url = {https://doi.org/10.1145/3757377.3763885},
doi = {10.1145/3757377.3763885},

booktitle = {Proceedings of the SIGGRAPH Asia 2025 Conference Papers},
articleno = {133},
numpages = {13},
location = {
},
series = {SA Conference Papers '25}
}

@article{he2022nemf,
  title={NeMF: Neural Motion Fields for Kinematic Animation},
  author={He, Chengan and Saito, Jun and Zachary, James and Rushmeier, Holly and Zhou, Yi},
  journal={Advances in Neural Information Processing Systems},
  volume={35},
  pages={4244--4256},
  year={2022}
}

@inproceedings{videopose3d,
  title={3D human pose estimation in video with temporal convolutions and semi-supervised training},
  author={Pavllo, Dario and Feichtenhofer, Christoph and Grangier, David and Auli, Michael},
  booktitle={Conference on Computer Vision and Pattern Recognition (CVPR)},
  year={2019}
}

@article{Chen2022LearningVM,
  title={Learning Variational Motion Prior for Video-based Motion Capture},
  author={Xin Chen and Zhuo Su and Lingbo Yang and Pei Cheng and Lan Xu and Bin Fu and Gang Yu},
  journal={ArXiv},
  year={2022},
  volume={abs/2210.15134},
  url={https://api.semanticscholar.org/CorpusID:253157307}
}

@inproceedings{
tevet2023human,
title={Human Motion Diffusion Model},
author={Guy Tevet and Sigal Raab and Brian Gordon and Yoni Shafir and Daniel Cohen-or and Amit Haim Bermano},
booktitle={The Eleventh International Conference on Learning Representations },
year={2023},
url={https://openreview.net/forum?id=SJ1kSyO2jwu}
}

@inproceedings{chen2023executing,
  title={Executing your Commands via Motion Diffusion in Latent Space},
  author={Chen, Xin and Jiang, Biao and Liu, Wen and Huang, Zilong and Fu, Bin and Chen, Tao and Yu, Gang},
  booktitle={Proceedings of the IEEE/CVF Conference on Computer Vision and Pattern Recognition},
  pages={18000--18010},
  year={2023}
}

@inproceedings{pinyoanuntapong2024mmm,
  title={MMM: Generative Masked Motion Model}, 
  author={Ekkasit Pinyoanuntapong and Pu Wang and Minwoo Lee and Chen Chen},
  booktitle={Proceedings of the IEEE/CVF Conference on Computer Vision and Pattern Recognition (CVPR)},
  year={2024},
}

@inproceedings{zhang2023generating,
  title={Generating human motion from textual descriptions with discrete representations},
  author={Zhang, Jianrong and Zhang, Yangsong and Cun, Xiaodong and Zhang, Yong and Zhao, Hongwei and Lu, Hongtao and Shen, Xi and Shan, Ying},
  booktitle={Proceedings of the IEEE/CVF conference on computer vision and pattern recognition},
  pages={14730--14740},
  year={2023}
}

@inproceedings{zhong2023attt2m,
  title={Attt2m: Text-driven human motion generation with multi-perspective attention mechanism},
  author={Zhong, Chongyang and Hu, Lei and Zhang, Zihao and Xia, Shihong},
  booktitle={Proceedings of the IEEE/CVF international conference on computer vision},
  pages={509--519},
  year={2023}
}

@misc{xie2025animamimicimitating3danimation,
      title={AnimaMimic: Imitating 3D Animation from Video Priors}, 
      author={Tianyi Xie and Yunuo Chen and Yaowei Guo and Yin Yang and Bolei Zhou and Demetri Terzopoulos and Ying Jiang and Chenfanfu Jiang},
      year={2025},
      eprint={2512.14133},
      archivePrefix={arXiv},
      primaryClass={cs.GR},
      url={https://arxiv.org/abs/2512.14133}, 
}

@inproceedings{anytop,
author = {Gat, Inbar and Raab, Sigal and Tevet, Guy and Reshef, Yuval and Bermano, Amit Haim and Cohen-Or, Daniel},
title = {AnyTop: Character Animation Diffusion with Any Topology},
year = {2025},
isbn = {9798400715402},
publisher = {Association for Computing Machinery},
address = {New York, NY, USA},
url = {https://doi.org/10.1145/3721238.3730621},
doi = {10.1145/3721238.3730621},
booktitle = {Proceedings of the Special Interest Group on Computer Graphics and Interactive Techniques Conference Conference Papers},
articleno = {13},
numpages = {10},
location = {
},
series = {SIGGRAPH Conference Papers '25}
}

@misc{mocapanything,
      title={MoCapAnything: Unified 3D Motion Capture for Arbitrary Skeletons from Monocular Videos}, 
      author={Kehong Gong and Zhengyu Wen and Weixia He and Mingxi Xu and Qi Wang and Ning Zhang and Zhengyu Li and Dongze Lian and Wei Zhao and Xiaoyu He and Mingyuan Zhang},
      year={2025},
      eprint={2512.10881},
      archivePrefix={arXiv},
      primaryClass={cs.CV},
      url={https://arxiv.org/abs/2512.10881}, 
}

@misc{li2025articulatedkinematicsdistillationvideo,
      title={Articulated Kinematics Distillation from Video Diffusion Models}, 
      author={Xuan Li and Qianli Ma and Tsung-Yi Lin and Yongxin Chen and Chenfanfu Jiang and Ming-Yu Liu and Donglai Xiang},
      year={2025},
      eprint={2504.01204},
      archivePrefix={arXiv},
      primaryClass={cs.GR},
      url={https://arxiv.org/abs/2504.01204}, 
}

@inproceedings{smplx,
author = {Pavlakos, Georgios and Choutas, Vasileios and Ghorbani, Nima and Bolkart, Timo and Osman, Ahmed and Tzionas, Dimitrios and Black, Michael},
year = {2019},
month = {06},
pages = {10967-10977},
title = {Expressive Body Capture: 3D Hands, Face, and Body From a Single Image},
doi = {10.1109/CVPR.2019.01123}
}

@inproceedings{bird,
author = {Kanazawa, Angjoo and Tulsiani, Shubham and Efros, Alexei A. and Malik, Jitendra},
title = {Learning Category-Specific Mesh Reconstruction from Image Collections},
year = {2018},
isbn = {978-3-030-01266-3},
publisher = {Springer-Verlag},
address = {Berlin, Heidelberg},
url = {https://doi.org/10.1007/978-3-030-01267-0_23},
doi = {10.1007/978-3-030-01267-0_23},
pages = {386–402},
numpages = {17},
location = {Munich, Germany}
}

@inproceedings{lassie,
author = {Yao, Chun-Han and Hung, Wei-Chih and Li, Yuanzhen and Rubinstein, Michael and Yang, Ming-Hsuan and Jampani, Varun},
title = {LASSIE: learning articulated shapes from sparse image ensemble via 3D part discovery},
year = {2022},
isbn = {9781713871088},
publisher = {Curran Associates Inc.},
address = {Red Hook, NY, USA},
booktitle = {Proceedings of the 36th International Conference on Neural Information Processing Systems},
articleno = {1113},
numpages = {13},
location = {New Orleans, LA, USA},
series = {NIPS '22}
}

@inproceedings {magicpony,
author = { Wu, Shangzhe and Li, Ruining and Jakab, Tomas and Rupprecht, Christian and Vedaldi, Andrea },
booktitle = { 2023 IEEE/CVF Conference on Computer Vision and Pattern Recognition (CVPR) },
title = {{ MagicPony: Learning Articulated 3D Animals in the Wild }},
year = {2023},
volume = {},
ISSN = {},
pages = {8792-8802},
doi = {10.1109/CVPR52729.2023.00849},
url = {https://doi.ieeecomputersociety.org/10.1109/CVPR52729.2023.00849},
publisher = {IEEE Computer Society},
address = {Los Alamitos, CA, USA},
month =Jun}

@inproceedings{smal,
  author={Zuffi, Silvia and Kanazawa, Angjoo and Jacobs, David W. and Black, Michael J.},
  booktitle={2017 IEEE Conference on Computer Vision and Pattern Recognition (CVPR)}, 
  title={3D Menagerie: Modeling the 3D Shape and Pose of Animals}, 
  year={2017},
  volume={},
  number={},
  pages={5524-5532},
  doi={10.1109/CVPR.2017.586}}

@inproceedings{mixermdm,
  author={Ruiz-Ponce, Pablo and Barquero, German and Palmero, Cristina and Escalera, Sergio and García-Rodríguez, José},
  booktitle={2025 IEEE/CVF Conference on Computer Vision and Pattern Recognition (CVPR)}, 
  title={MixerMDM: Learnable Composition of Human Motion Diffusion Models}, 
  year={2025},
  volume={},
  number={},
  pages={12380-12390},
  doi={10.1109/CVPR52734.2025.01155}}

@inproceedings{lgtm,
author = {Sun, Haowen and Zheng, Ruikun and Huang, Haibin and Ma, Chongyang and Huang, Hui and Hu, Ruizhen},
title = {LGTM: Local-to-Global Text-Driven Human Motion Diffusion Model},
year = {2024},
isbn = {9798400705250},
publisher = {Association for Computing Machinery},
address = {New York, NY, USA},
url = {https://doi.org/10.1145/3641519.3657422},
doi = {10.1145/3641519.3657422},
booktitle = {ACM SIGGRAPH 2024 Conference Papers},
articleno = {66},
numpages = {9},
location = {Denver, CO, USA},
series = {SIGGRAPH '24}
}

@inproceedings{energymogen,
author = {Zhang, Jianrong and Fan, Hehe and Yang, Yi},
year = {2025},
month = {06},
pages = {17592-17602},
title = {EnergyMogen: Compositional Human Motion Generation with Energy-Based Diffusion Model in Latent Space},
doi = {10.1109/CVPR52734.2025.01639}
}

@misc{gpt52,
  author       = {{OpenAI}},
  title        = {Update to GPT-5 System Card: GPT-5.2},
  year         = {2025},
  month        = dec,
  howpublished = {\url{https://cdn.openai.com/pdf/3a4153c8-c748-4b71-8e31-aecbde944f8d/oai_5_2_system-card.pdf}},
  note         = {Accessed: 2026-01-22},
}

@misc{truebones,
  title        = {Truebones Motion Capture},
  author       = {{Truebones}},
  howpublished = {\url{https://truebones.gumroad.com/}},
  note         = {Accessed: 2025-12-01},
  year         = {n.d.}
}

@conference{amass,
  title = {{AMASS}: Archive of Motion Capture as Surface Shapes},
  author = {Mahmood, Naureen and Ghorbani, Nima and Troje, Nikolaus F. and Pons-Moll, Gerard and Black, Michael J.},
  booktitle = {International Conference on Computer Vision},
  pages = {5442--5451},
  month = oct,
  year = {2019},
  month_numeric = {10}
}

@inproceedings{humanml3d,
    author    = {Guo, Chuan and Zou, Shihao and Zuo, Xinxin and Wang, Sen and Ji, Wei and Li, Xingyu and Cheng, Li},
    title     = {Generating Diverse and Natural 3D Human Motions From Text},
    booktitle = {Proceedings of the IEEE/CVF Conference on Computer Vision and Pattern Recognition (CVPR)},
    month     = {June},
    year      = {2022},
    pages     = {5152-5161}
}

@article{lafan,
author    = {Félix G. Harvey and Mike Yurick and Derek Nowrouzezahrai and Christopher Pal},
title     = {Robust Motion In-Betweening},
booktitle = {ACM Transactions on Graphics (Proceedings of ACM SIGGRAPH)},
publisher = {ACM},
volume    = {39},
number    = {4},
year      = {2020}
}

@inproceedings{human36m,
    author    = {Zhu, Yue and Samet, Nermin and Picard, David},
    title     = {H3WB: Human3.6M 3D WholeBody Dataset and Benchmark},
    booktitle = {Proceedings of the IEEE/CVF International Conference on Computer Vision (ICCV)},
    month     = {October},
    year      = {2023},
    pages     = {20166-20177}
}

@article{motionx,
  title={Motion-X++: A Large-Scale Multimodal 3D Whole-body Human Motion Dataset},
  author={Zhang, Yuhong and Lin, Jing and Zeng, Ailing and Wu, Guanlin and Lu, Shunlin and Fu, Yurong and Cai, Yuanhao and Zhang, Ruimao and Wang, Haoqian and Zhang, Lei},
  journal={arXiv preprint arXiv:2501.05098},
  year={2025}
}

@inproceedings{zoo300k,
  author       = {Zeyu Zhang and
                  Yiran Wang and
                  Biao Wu and
                  Shuo Chen and
                  Zhiyuan Zhang and
                  Shiya Huang and
                  Wenbo Zhang and
                  Meng Fang and
                  Ling Chen and
                  Yang Zhao},
  title        = {Motion Avatar: Generate Human and Animal Avatars with Arbitrary Motion},
  booktitle    = {35th British Machine Vision Conference, {BMVC} 2024, Glasgow, UK,
                  November 25-28, 2024},
  publisher    = {{BMVA} Press},
  year         = {2024},
  url          = {https://bmvc2024.org/proceedings/185/},
  bibsource    = {dblp computer science bibliography, https://dblp.org}
}

@inproceedings{objaverseXL,
  author       = {Matt Deitke and
                  Ruoshi Liu and
                  Matthew Wallingford and
                  Huong Ngo and
                  Oscar Michel and
                  Aditya Kusupati and
                  Alan Fan and
                  Christian Laforte and
                  Vikram Voleti and
                  Samir Yitzhak Gadre and
                  Eli VanderBilt and
                  Aniruddha Kembhavi and
                  Carl Vondrick and
                  Georgia Gkioxari and
                  Kiana Ehsani and
                  Ludwig Schmidt and
                  Ali Farhadi},
  editor       = {Alice Oh and
                  Tristan Naumann and
                  Amir Globerson and
                  Kate Saenko and
                  Moritz Hardt and
                  Sergey Levine},
  title        = {Objaverse-XL: {A} Universe of 10M+ 3D Objects},
  booktitle    = {Advances in Neural Information Processing Systems 36: Annual Conference
                  on Neural Information Processing Systems 2023, NeurIPS 2023, New Orleans,
                  LA, USA, December 10 - 16, 2023},
  year         = {2023},
  url          = {http://papers.nips.cc/paper\_files/paper/2023/hash/70364304877b5e767de4e9a2a511be0c-Abstract-Datasets\_and\_Benchmarks.html},
  bibsource    = {dblp computer science bibliography, https://dblp.org}
}

@article{fusion4d,
author = {Dou, Mingsong and Khamis, Sameh and Degtyarev, Yury and Davidson, Philip and Fanello, Sean Ryan and Kowdle, Adarsh and Escolano, Sergio Orts and Rhemann, Christoph and Kim, David and Taylor, Jonathan and Kohli, Pushmeet and Tankovich, Vladimir and Izadi, Shahram},
title = {Fusion4D: real-time performance capture of challenging scenes},
year = {2016},
issue_date = {July 2016},
publisher = {Association for Computing Machinery},
address = {New York, NY, USA},
volume = {35},
number = {4},
issn = {0730-0301},
url = {https://doi.org/10.1145/2897824.2925969},
doi = {10.1145/2897824.2925969},
journal = {ACM Trans. Graph.},
month = jul,
articleno = {114},
numpages = {13}
}

@article{sohn2015learning,
  title={Learning structured output representation using deep conditional generative models},
  author={Sohn, Kihyuk and Lee, Honglak and Yan, Xinchen},
  journal={Advances in neural information processing systems},
  volume={28},
  year={2015}
}

@article{zhang20233dshape2vecset,
  title={3dshape2vecset: A 3d shape representation for neural fields and generative diffusion models},
  author={Zhang, Biao and Tang, Jiapeng and Niessner, Matthias and Wonka, Peter},
  journal={ACM Transactions on Graphics (TOG)},
  volume={42},
  number={4},
  pages={1--16},
  year={2023},
  publisher={ACM New York, NY, USA}
}

@article{attn,
  title={Attention is all you need},
  author={Vaswani, A},
  journal={Advances in Neural Information Processing Systems},
  year={2017}
}

@inproceedings{zeng2022smoothnet,
      title={SmoothNet: A Plug-and-Play Network for Refining Human Poses in Videos},
      author={Zeng, Ailing and Yang, Lei and Ju, Xuan and Li, Jiefeng and Wang, Jianyi and Xu, Qiang},
      booktitle={European Conference on Computer Vision},
      year={2022},
      organization={Springer}
}

@inproceedings{lipman2022flow,
  author       = {Yaron Lipman and
                  Ricky T. Q. Chen and
                  Heli Ben{-}Hamu and
                  Maximilian Nickel and
                  Matthew Le},
  title        = {Flow Matching for Generative Modeling},
  booktitle    = {The Eleventh International Conference on Learning Representations,
                  {ICLR} 2023, Kigali, Rwanda, May 1-5, 2023},
  publisher    = {OpenReview.net},
  year         = {2023},
  url          = {https://openreview.net/forum?id=PqvMRDCJT9t},
  bibsource    = {dblp computer science bibliography, https://dblp.org}
}

@misc{simeoni2025dinov3,
  title={{DINOv3}},
  author={Sim{\'e}oni, Oriane and Vo, Huy V. and Seitzer, Maximilian and Baldassarre, Federico and Oquab, Maxime and Jose, Cijo and Khalidov, Vasil and Szafraniec, Marc and Yi, Seungeun and Ramamonjisoa, Micha{\"e}l and Massa, Francisco and Haziza, Daniel and Wehrstedt, Luca and Wang, Jianyuan and Darcet, Timoth{\'e}e and Moutakanni, Th{\'e}o and Sentana, Leonel and Roberts, Claire and Vedaldi, Andrea and Tolan, Jamie and Brandt, John and Couprie, Camille and Mairal, Julien and J{\'e}gou, Herv{\'e} and Labatut, Patrick and Bojanowski, Piotr},
  year={2025},
  eprint={2508.10104},
  archivePrefix={arXiv},
  primaryClass={cs.CV},
  url={https://arxiv.org/abs/2508.10104},
}

@misc{hunyuan3d,
      title={Hunyuan3D 2.1: From Images to High-Fidelity 3D Assets with Production-Ready PBR Material}, 
      author={Team Hunyuan3D and Shuhui Yang and Mingxin Yang and Yifei Feng and Xin Huang and Sheng Zhang and Zebin He and Di Luo and Haolin Liu and Yunfei Zhao and Qingxiang Lin and Zeqiang Lai and Xianghui Yang and Huiwen Shi and Zibo Zhao and Bowen Zhang and Hongyu Yan and Lifu Wang and Sicong Liu and Jihong Zhang and Meng Chen and Liang Dong and Yiwen Jia and Yulin Cai and Jiaao Yu and Yixuan Tang and Dongyuan Guo and Junlin Yu and Hao Zhang and Zheng Ye and Peng He and Runzhou Wu and Shida Wei and Chao Zhang and Yonghao Tan and Yifu Sun and Lin Niu and Shirui Huang and Bojian Zheng and Shu Liu and Shilin Chen and Xiang Yuan and Xiaofeng Yang and Kai Liu and Jianchen Zhu and Peng Chen and Tian Liu and Di Wang and Yuhong Liu and Linus and Jie Jiang and Jingwei Huang and Chunchao Guo},
      year={2025},
      eprint={2506.15442},
      archivePrefix={arXiv},
      primaryClass={cs.CV},
      url={https://arxiv.org/abs/2506.15442}, 
}

@misc{triposg,
      title={TripoSG: High-Fidelity 3D Shape Synthesis using Large-Scale Rectified Flow Models}, 
      author={Yangguang Li and Zi-Xin Zou and Zexiang Liu and Dehu Wang and Yuan Liang and Zhipeng Yu and Xingchao Liu and Yuan-Chen Guo and Ding Liang and Wanli Ouyang and Yan-Pei Cao},
      year={2025},
      eprint={2502.06608},
      archivePrefix={arXiv},
      primaryClass={cs.CV},
      url={https://arxiv.org/abs/2502.06608}, 
}

@inproceedings{genzoo,
  author    = {Niewiadomski, Tomasz and Yiannakidis, Anastasios and Cuevas-Velasquez, Hanz and Sanyal, Soubhik and Black, Michael J. and Zuffi, Silvia and Kulits, Peter},
  title     = {Generative Zoo},
  booktitle = {Proceedings of the IEEE/CVF International Conference on Computer Vision (ICCV)},
  year      = {2025}
}

@misc{fan2025zerozeroshotmotiongeneration,
      title={Go to Zero: Towards Zero-shot Motion Generation with Million-scale Data}, 
      author={Ke Fan and Shunlin Lu and Minyue Dai and Runyi Yu and Lixing Xiao and Zhiyang Dou and Junting Dong and Lizhuang Ma and Jingbo Wang},
      year={2025},
      eprint={2507.07095},
      archivePrefix={arXiv},
      primaryClass={cs.CV},
      url={https://arxiv.org/abs/2507.07095}, 
}

@inproceedings{guo2024momask,
  title={Momask: Generative masked modeling of 3d human motions},
  author={Guo, Chuan and Mu, Yuxuan and Javed, Muhammad Gohar and Wang, Sen and Cheng, Li},
  booktitle={Proceedings of the IEEE/CVF Conference on Computer Vision and Pattern Recognition},
  pages={1900--1910},
  year={2024}
}

@inproceedings{zhang2025towards,
  author = {Zhang, Zongye and Kong, Bohan and Liu, Qingjie and Wang, Yunhong},
  title = {Towards Robust and Controllable Text-to-Motion via Masked Autoregressive Diffusion},
  year = {2025},
  isbn = {9798400720352},
  publisher = {Association for Computing Machinery},
  address = {New York, NY, USA},
  url = {https://doi.org/10.1145/3746027.3754748},
  doi = {10.1145/3746027.3754748},
  booktitle = {Proceedings of the 33rd ACM International Conference on Multimedia},
  pages = {9326–9335},
  numpages = {10},
  location = {Dublin, Ireland},
  series = {MM '25}
}

@article{xu2020rignet,
  author       = {Zhan Xu and
                  Yang Zhou and
                  Evangelos Kalogerakis and
                  Chris Landreth and
                  Karan Singh},
  title        = {RigNet: neural rigging for articulated characters},
  journal      = {{ACM} Trans. Graph.},
  volume       = {39},
  number       = {4},
  pages        = {58},
  year         = {2020},
  url          = {https://doi.org/10.1145/3386569.3392379},
  doi          = {10.1145/3386569.3392379},
  bibsource    = {dblp computer science bibliography, https://dblp.org}
}

@article{wan2025,
      title={Wan: Open and Advanced Large-Scale Video Generative Models}, 
      author={Team Wan and Ang Wang and Baole Ai and Bin Wen and Chaojie Mao and Chen-Wei Xie and Di Chen and Feiwu Yu and Haiming Zhao and Jianxiao Yang and Jianyuan Zeng and Jiayu Wang and Jingfeng Zhang and Jingren Zhou and Jinkai Wang and Jixuan Chen and Kai Zhu and Kang Zhao and Keyu Yan and Lianghua Huang and Mengyang Feng and Ningyi Zhang and Pandeng Li and Pingyu Wu and Ruihang Chu and Ruili Feng and Shiwei Zhang and Siyang Sun and Tao Fang and Tianxing Wang and Tianyi Gui and Tingyu Weng and Tong Shen and Wei Lin and Wei Wang and Wei Wang and Wenmeng Zhou and Wente Wang and Wenting Shen and Wenyuan Yu and Xianzhong Shi and Xiaoming Huang and Xin Xu and Yan Kou and Yangyu Lv and Yifei Li and Yijing Liu and Yiming Wang and Yingya Zhang and Yitong Huang and Yong Li and You Wu and Yu Liu and Yulin Pan and Yun Zheng and Yuntao Hong and Yupeng Shi and Yutong Feng and Zeyinzi Jiang and Zhen Han and Zhi-Fan Wu and Ziyu Liu},
      journal = {arXiv preprint arXiv:2503.20314},
      year={2025}
}

@inproceedings{jaegle2021perceiverio,
  author       = {Andrew Jaegle and
                  Sebastian Borgeaud and
                  Jean{-}Baptiste Alayrac and
                  Carl Doersch and
                  Catalin Ionescu and
                  David Ding and
                  Skanda Koppula and
                  Daniel Zoran and
                  Andrew Brock and
                  Evan Shelhamer and
                  Olivier J. H{\'{e}}naff and
                  Matthew M. Botvinick and
                  Andrew Zisserman and
                  Oriol Vinyals and
                  Jo{\~{a}}o Carreira},
  title        = {Perceiver {IO:} {A} General Architecture for Structured Inputs {\&}
                  Outputs},
  booktitle    = {The Tenth International Conference on Learning Representations, {ICLR}
                  2022, Virtual Event, April 25-29, 2022},
  publisher    = {OpenReview.net},
  year         = {2022},
  url          = {https://openreview.net/forum?id=fILj7WpI-g},
  bibsource    = {dblp computer science bibliography, https://dblp.org}
}

@article{
  velickovic2018graph,
  title="{Graph Attention Networks}",
  author={Veli{\v{c}}kovi{\'{c}}, Petar and Cucurull, Guillem and Casanova, Arantxa and Romero, Adriana and Li{\`{o}}, Pietro and Bengio, Yoshua},
  journal={International Conference on Learning Representations},
  year={2018},
  url={https://openreview.net/forum?id=rJXMpikCZ},
  note={accepted as poster},
}

@article{peng2024charactergen,
  title   ={CharacterGen: Efficient 3D Character Generation from Single Images with Multi-View Pose Canonicalization}, 
  author  ={Hao-Yang Peng and Jia-Peng Zhang and Meng-Hao Guo and Yan-Pei Cao and Shi-Min Hu},
  journal ={ACM Transactions on Graphics (TOG)},
  year    ={2024},
  volume  ={43},
  number  ={4}
}

@inproceedings{lxl,
  author       = {Xiao{-}Lei Li and
                  Hao{-}Xiang Chen and
                  Yanni Zhang and
                  Kai Ma and
                  Alan Zhao and
                  Tai{-}Jiang Mu and
                  Hao{-}Xiang Guo and
                  Ran Zhang},
  editor       = {Ginger Alford and
                  Hao (Richard) Zhang and
                  Adriana Schulz},
  title        = {{RELATE3D:} REfocusing Latent Adapter for Targeted local Enhancement
                  and Editing in 3D Generation},
  booktitle    = {Proceedings of the Special Interest Group on Computer Graphics and
                  Interactive Techniques Conference, {SIGGRAPH} Conference Papers 2025,
                  Vancouver, BC, Canada, August 10-14, 2025},
  pages        = {79:1--79:12},
  publisher    = {{ACM}},
  year         = {2025},
  url          = {https://doi.org/10.1145/3721238.3730648},
  doi          = {10.1145/3721238.3730648},
  bibsource    = {dblp computer science bibliography, https://dblp.org}
}

@article{Hou2024, 
author = {Shuaiying Hou and Congyi Wang and Wenlin Zhuang and Yu Chen and Yangang Wang and Hujun Bao and Jinxiang Chai and Weiwei Xu},
title = {A causal convolutional neural network for multi-subject motion modeling and generation},
year = {2024},
journal = {Computational Visual Media},
volume = {10},
number = {1},
pages = {45-59},
url = {https://www.sciopen.com/article/10.1007/s41095-022-0307-3},
doi = {10.1007/s41095-022-0307-3},
}

@article{diffusionmodels,
  author={Wang, Chen and Peng, Hao-Yang and Liu, Ying-Tian and Gu, Jiatao and Hu, Shi-Min},
  journal={Computational Visual Media}, 
  title={Diffusion Models for 3D Generation: A Survey}, 
  year={2025},
  volume={11},
  number={1},
  pages={1-28},
  doi={10.26599/CVM.2025.9450452}}

@article{deepmotionretarget,
author = {Aberman, Kfir and Li, Peizhuo and Lischinski, Dani and Sorkine-Hornung, Olga and Cohen-Or, Daniel and Chen, Baoquan},
title = {Skeleton-aware networks for deep motion retargeting},
year = {2020},
issue_date = {August 2020},
publisher = {Association for Computing Machinery},
address = {New York, NY, USA},
volume = {39},
number = {4},
issn = {0730-0301},
url = {https://doi.org/10.1145/3386569.3392462},
doi = {10.1145/3386569.3392462},
journal = {ACM Trans. Graph.},
month = aug,
articleno = {62},
numpages = {14}
}

@inproceedings{posetomotion,
author = {Zhao, Qingqing and Li, Peizhuo and Yifan, Wang and Olga, Sorkine-Hornung and Wetzstein, Gordon},
title = {Pose-to-Motion: Cross-Domain Motion Retargeting with Pose Prior},
year = {2024},
publisher = {Eurographics Association},
address = {Goslar, DEU},
url = {https://doi.org/10.1111/cgf.15170},
doi = {10.1111/cgf.15170},
booktitle = {Proceedings of the ACM SIGGRAPH/Eurographics Symposium on Computer Animation},
pages = {1–10},
numpages = {10},
location = {Montreal, Quebec, Canada},
series = {SCA '24}
}

@inproceedings{motion2motion,
author = {Chen, Ling-Hao and Zhang, Yuhong and Yin, Zixin and Dou, Zhiyang and Chen, Xin and Wang, Jingbo and Komura, Taku and Zhang, Lei},
title = {Motion2Motion: Cross-topology Motion Transfer with Sparse Correspondence},
year = {2025},
isbn = {9798400721373},
publisher = {Association for Computing Machinery},
address = {New York, NY, USA},
url = {https://doi.org/10.1145/3757377.3763811},
doi = {10.1145/3757377.3763811},
booktitle = {Proceedings of the SIGGRAPH Asia 2025 Conference Papers},
articleno = {150},
numpages = {11},
location = {
},
series = {SA Conference Papers '25}
}
